\documentclass{article}

\usepackage{microtype}
\usepackage{graphicx}
\usepackage{subcaption}
\usepackage{booktabs} 

\usepackage{hyperref}

\usepackage{algorithm}        
\usepackage{algorithmic}    

\usepackage[preprint]{icml2026}

\usepackage{amsmath}
\usepackage{amssymb}
\usepackage{mathtools}
\usepackage{amsthm}
\usepackage{tikz}
\usetikzlibrary{automata,positioning}

\usepackage[capitalize,noabbrev]{cleveref}

\usepackage{comment}
\usepackage[dvipsnames]{xcolor}
\usepackage[nolist]{acronym}
\begin{acronym}[DRAM]
    \acro{AI}{Artificial Intelligence}
    \acrodefindefinite{AI}{an}{an}
    \acro{ASIC}{application-specific integrated circuit}
    \acrodefindefinite{ASIC}{an}{an}
    \acro{BD}{bounded delay}
    \acro{NN}{Neural Network}
    \acro{BNN}{Binarized Neural Network}
    \acro{CNN}{Convolutional Neural Network}
    \acro{CTM}{Convolutional Tsetlin machine}
    \acro{CoTM}{Coalesced Tsetlin Machine}
    \acro{FSM}{Finite state machine}
    \acro{LA}{learning automaton}
    \acrodefplural{LA}{learning automata}
    \acrodefindefinite{LA}{an}{a}
    \acro{ML}{Machine Learning}
    \acrodefindefinite{ML}{an}{a}
    \acro{TA}{Tsetlin automaton}
    \acrodefplural{TA}{Tsetlin automata}
    \acro{TAT}{Tsetlin automaton team}
    \acro{TM}{Tsetlin Machine}
    \acro{GraphTM}{Graph Tsetlin Machine}
    \acro{GCN}{Graph Convolutional Neural Network}
    \acro{GNN}{Graph Neural Network}
\end{acronym}
\usepackage{bbold}
\usepackage{multirow}
\usepackage{pdflscape}
\usepackage{xurl}

\newcommand{\True}{1}   
\newcommand{\False}{0}  

\theoremstyle{plain}

\theoremstyle{definition}

\theoremstyle{remark}

\usepackage[textsize=tiny]{todonotes}

\icmltitlerunning{The Tsetlin Machine Goes Deep: Logical Learning and Reasoning With Graphs}

\begin{document}

\twocolumn[
  \icmltitle{The Tsetlin Machine Goes Deep: Logical Learning and Reasoning With Graphs}



  \icmlsetsymbol{equal}{*}

  \begin{icmlauthorlist}
    \icmlauthor{Ole-Christoffer Granmo}{uia}
    \icmlauthor{Youmna Abdelwahab}{equal,uia}
    \icmlauthor{Per-Arne Andersen}{equal,uia}
    \icmlauthor{Karl Audun K. Borgersen}{equal,uia}
    \icmlauthor{Paul F. A. Clarke}{equal,uia}
    \icmlauthor{Kunal Dumbre}{equal,uia}
    \icmlauthor{Ylva Grønningsæter}{equal,uia}
    \icmlauthor{Vojtech Halenka}{equal,uia}
    \icmlauthor{Runar Helin}{equal,uia}
    \icmlauthor{equal,Lei Jiao}{uia}\icmlauthor{Ahmed Khalid}{equal,uia}
    \icmlauthor{Rebekka Omslandseter}{equal,uia}
    \icmlauthor{Rupsa Saha}{equal,uia}
    \icmlauthor{Mayur Shende}{equal,uia}
     \icmlauthor{Xuan Zhang}{equal,uia}
  \end{icmlauthorlist}

  \icmlaffiliation{uia}{Centre for AI Research, University of Agder, Norway}
  \icmlcorrespondingauthor{Xuan Zhang}{xuan.zhang@uia.no}

  \icmlkeywords{Machine Learning, ICML}

  \vskip 0.3in
]



\printAffiliationsAndNotice{}  

\begin{abstract}
  Pattern recognition with concise and flat AND-rules makes the \ac{TM} both interpretable and efficient, while the power of Tsetlin automata enables accuracy comparable to deep learning on an increasing number of datasets. We introduce the \ac{GraphTM} for learning \emph{interpretable deep clauses} from \emph{graph-structured} input. Moving beyond flat, fixed-length input, the \ac{GraphTM} gets more versatile, supporting sequences, grids, relations, and multimodality. Through message passing, the \ac{GraphTM} builds nested deep clauses to recognize sub-graph patterns with exponentially fewer clauses, increasing both interpretability and data utilization. For image classification, \ac{GraphTM} preserves interpretability and achieves $3.86\%$-points higher accuracy on CIFAR-10 than a convolutional TM. For tracking action coreference, faced with increasingly challenging tasks, \ac{GraphTM} outperforms other reinforcement learning methods by up to $20.6\%$-points. In recommendation systems, it tolerates increasing noise to a great extent similar to a GCN. Finally, for viral genome sequence data, \ac{GraphTM} is competitive with BiLSTM-CNN and \ac{GCN} accuracy-wise, training $\sim2.5\times$ faster than \ac{GCN}. The \ac{GraphTM}'s application to these varied fields demonstrates how graph representation learning and deep clauses bring new possibilities for TM learning. 
  
\end{abstract}

\section{Introduction}
\label{intro}

Recent literature highlights the pattern recognition power of \acp{TM} using interpretable clauses, providing accuracy competitive with deep learning approaches~\citep{jeeru2025jamming, saha_using_2023, yadav-etal-2021-enhancing, bhattarai-etal-2024-tsetlin}. However, its Boolean representation of input data hinders its widespread application. Recently, the use of hypervectors~\citep{halenka2024exploring} has significantly expanded the types of data that can be processed by a \ac{TM}, while still being constrained to standard \ac{TM} architectures. We here propose the \ac{GraphTM}, which can process multimodal data represented as hypervectorized directed and labeled multigraphs. This enhancement ensures that a \ac{TM}-based system is no longer restricted to fixed-sized or monolithic input. While graphs have been extensively used to represent complex structures involving multimodal data in the context of \acp{GNN}~\citep{ektefaie_multimodal_2023}, no such mechanism has been available for \ac{TM}s prior to this work. The \ac{GraphTM} is an advancement of the Hypervector \ac{TM}, and consequently on the standard \ac{TM}~\citep{granmo2018tsetlin}. The \ac{CTM}~\citep{granmo2019convolutional} is built upon the strengths of a standard \ac{TM} to allow immediate context (within the convolutional window) to be captured by the clauses. The \ac{GraphTM} enhances this capability by allowing context to be gathered over the graph topography, as opposed to the strictly physical locality seen in \acp{CTM}.

\begin{figure*}[htbp]
    \centering
    \includegraphics[width=0.57\linewidth]{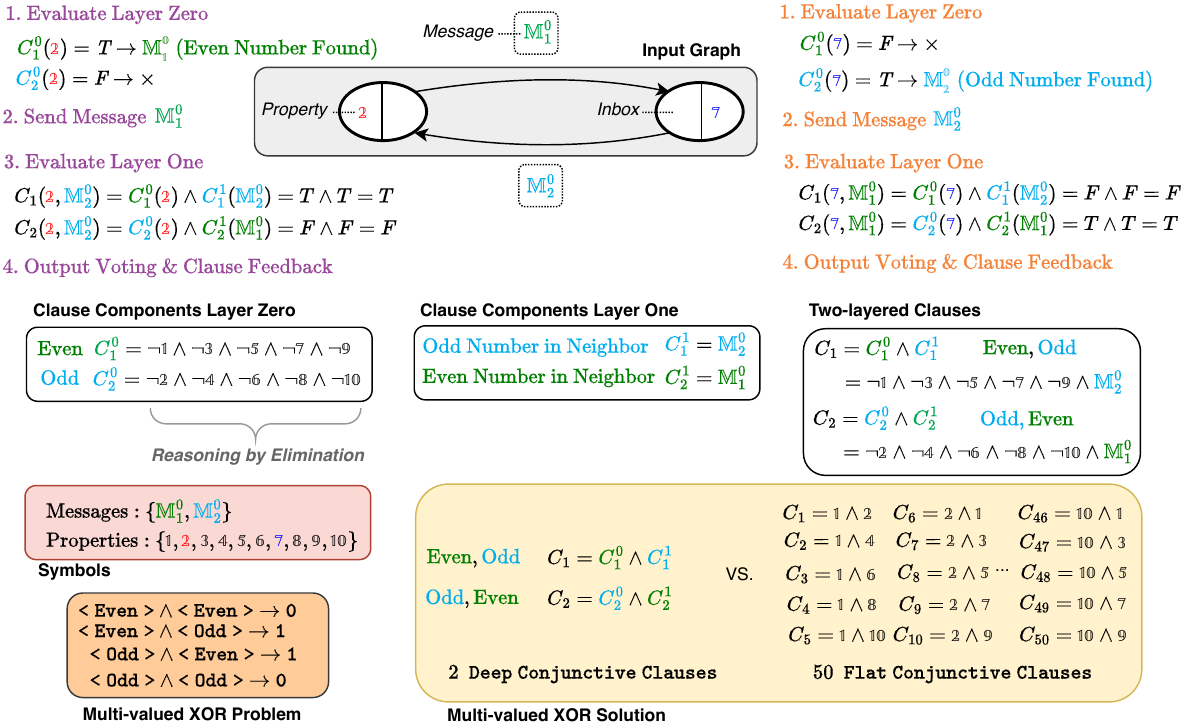}
    \caption{The \ac{GraphTM} processes graph-structured input and exploits this structure to build \emph{deep} clauses through \emph{nesting}. \emph{Reasoning by elimination} reduces the number of clauses \emph{exponentially}, while the processing of graph nodes and clauses is \emph{parallel} (\textcolor{Purple}{purple} and \textcolor{Orange}{orange}). The \ac{GraphTM} first evaluates each node's properties using the layer-zero components $C_j^0$ of the clauses (1.). If $C_j^0$ matches the properties of a node, the clause signals this to neighboring nodes by sending a message $\boldsymbol{M}_j^0$ (2.). Upon receiving messages in their inboxes, the nodes evaluate the layer-one clause components $C^1_j$, which gives the value of the full clause $C_j$ for a two-layer system (3.). Finally, the classification and clause updating follow the standard \ac{TM} approach (4.). Both the node properties and messages are \emph{symbols} in \emph{hypervector space}, going beyond the Boolean representation of \acp{TM}, yet maintaining \emph{interpretability.}}
    \label{fig:graph_tsetlin_machine_conceptual}
\end{figure*}

\section{Graph Tsetlin Machine}
\label{sec:gtm}

The ability to learn patterns from graph-structured input is central to the \ac{GraphTM}, to allow for classification~\citep{dropclause}, regression~\citep{abeyrathna2020nonlinear}, auto-encoding~\citep{bhattarai-etal-2024-tsetlin}, or contextual bandit learning~\citep{Raihan2022contextual}. In the following, we use the so-called Multi-Valued XOR Problem in Figure~\ref{fig:graph_tsetlin_machine_conceptual} to describe the GraphTM approach step-by-step (see Appendix \ref{app:GraphTM} for a more rigorous exposition). 

\textbf{Graph Input.} The \ac{GraphTM} takes as input a multigraph $G=(V, E, \boldsymbol{P}, \boldsymbol{T})$ with nodes (vertices) $v_q \in V$ and typed edges $(v_q, v_r, \boldsymbol{t}) \in E$. Each node $v_q$ has properties $\boldsymbol{p}_k \in \boldsymbol{P}_q \subseteq \boldsymbol{P}$ taken from a set of properties $\boldsymbol{P}$. The type of an edge $\boldsymbol{t} \in \boldsymbol{T}$ is selected from the set of available edge types $\boldsymbol{T}$. For example, Figure~\ref{fig:graph_tsetlin_machine_conceptual} shows an input graph of two interconnected nodes. The available properties are  $\boldsymbol{P} = \{\boldsymbol{1}, \textcolor{red}{\boldsymbol{2}}, \boldsymbol{3}, \boldsymbol{4}, \boldsymbol{5}, \boldsymbol{6}, \textcolor{blue}{\boldsymbol{7}}, \boldsymbol{8}, \boldsymbol{9}, \boldsymbol{10}\}$ while there is only one edge type $\boldsymbol{T} = \{\mathit{Plain}\}$. The left node has property~$\textcolor{red}{\boldsymbol{2}}$  while the right node has property~$\textcolor{blue}{\boldsymbol{7}}$. Note that a property can represent any type of data, for instance a pixel in image processing.

\textbf{Clauses.} A vanilla \ac{TM} builds conjunctive clauses to recognize patterns from propositional features $X = [x_1, x_2, \ldots, x_o]$. A clause is flat, taking the form $C_j = \bigwedge_{{{l_k}} {\in} L_j} l_k$, where $L_j$ is a subset of the features and their negation, referred to as literals: $L_j \subseteq \{x_1, x_2 \ldots, x_o, \lnot x_1, \lnot x_2, \ldots, \lnot x_o\}$ (see Appendices \ref{app:TM} and \ref{app:CoTM} for a full description of \ac{TM}). The \ac{GraphTM} clauses, on the other hand, operate on properties $\boldsymbol{P}$. A clause literal $\boldsymbol{p}_k$/$\lnot \boldsymbol{p}_k$ then specifies the \emph{presence}/\emph{absence} of property $\boldsymbol{p}_k$ in a node. The clause $C_j = \boldsymbol{1} \land \boldsymbol{2}$, for example, says that the properties $\boldsymbol{1}$ and $\boldsymbol{2}$ must both be present for the clause to be \emph{True}.

\textbf{Deep Clauses.} A \ac{GraphTM} clause further subdivides into components to create a \emph{deep} clause, one component per layer: $C_j = C_j^0 \land C_j^1 \land \cdots \land C_j^{D-1}$, with $D$ being the layer index (depth). The layer-zero components $C_j^0$ operate on the node properties $\boldsymbol{P}$ as described above. In Figure~\ref{fig:graph_tsetlin_machine_conceptual}, for example, the clause component $\textcolor{Green}{C_1^0} = \lnot 1 \land \lnot 3 \land \lnot 5 \land \lnot 7 \land \lnot 9$ specifies the absence of odd numbers in a node, matching even numbers through \emph{reasoning by elimination}.

\textbf{Messages.} A set of message symbols $\boldsymbol{M}$ connects the layers of a \ac{GraphTM}. In Figure~\ref{fig:graph_tsetlin_machine_conceptual}, we have two kinds of messages: $\boldsymbol{M} = \{ \textcolor{Green}{\boldsymbol{M}_1^0}, \textcolor{Cyan}{\boldsymbol{M}_2^0}\}$. Each component $C_j^0$ uses a dedicated message $\boldsymbol{M}_j^0 \in \boldsymbol{M}$ to signal that it is \emph{True} at a node. Further, each node $v_q$ in the graph gets an \emph{inbox} $I_q^d$ for storing messages from layer $d$. The clause components $C_j^i$ of the subsequent layers $i > 0$ then check the inbox for messages $\boldsymbol{M}_j^{i-1}$ from the previous layer, to produce messages $\boldsymbol{M}_j^i$ for the current layer. The clause component $\textcolor{Green}{C_1^1} = \textcolor{Green}{\boldsymbol{M}_1^0}$ for layer one in Figure~\ref{fig:graph_tsetlin_machine_conceptual}, for instance, specifies that the message $\textcolor{Green}{\boldsymbol{M}_1^0}$ must be in the inbox.  

\textbf{Message Submission.} Every time a clause component produces a message at a node, as described above, it sends the message to the inboxes of the node's neighbors according to the graph edges $E$. Figure~\ref{fig:graph_tsetlin_machine_conceptual} illustrates how the clause layer zero component $\textcolor{Green}{C_1^0}$ matches the property $\textcolor{red}{\boldsymbol{2}}$ of the left node, and thus sends the message $\textcolor{Green}{\boldsymbol{M}_1^0}$ to the inbox of the right node. Upon receiving the message, the clause component $\textcolor{Green}{C_2^1} = \textcolor{Green}{\boldsymbol{M}_1^0}$ of layer one can be evaluated (i.e., by comparing it with the corresponding message symbols to determine whether they match) to obtain the truth value of the complete clause $C_2 = \textcolor{Cyan}{C_2^0} \land \textcolor{Green}{C_2^1}$ for the right node.  In this manner, the nodes and clauses of each layer are processed in parallel, using one thread per clause-node pair.

\textbf{Edge Types.} When the graph has multiple edge types $\boldsymbol{t}^t_{qr} \in \boldsymbol{T}$, the messages are annotated with the type of edge they pass along. This information is stored in the inbox of a node along with the messages, for additional context.

\textbf{Vector Symbolic Computation.} Internally, the \ac{GraphTM} uses sparse hypervectors~\citep{rachkovskij2001representation} to encode node properties, messages, and edge types. They are all symbols $\boldsymbol{S}$ from a hypervector perspective: $\boldsymbol{S} = \boldsymbol{P} \cup \boldsymbol{M} \cup \boldsymbol{T}$. The hypervectors are in Boolean form, supporting \ac{TM}-based learning and reasoning~\citep{halenka2024exploring}. For our example in Figure~\ref{fig:graph_tsetlin_machine_conceptual}, we thus have the symbols $\boldsymbol{S} = \{ \boldsymbol{1}, \textcolor{red}{\boldsymbol{2}}, \boldsymbol{3}, \boldsymbol{4}, \boldsymbol{5}, \boldsymbol{6}, \textcolor{blue}{\boldsymbol{7}}, \boldsymbol{8}, \boldsymbol{9}, \boldsymbol{10}, \textcolor{Green}{\boldsymbol{M}_1^0}, \textcolor{Cyan}{\boldsymbol{M}_2^0}\}$ (no edge types). We use the bundle operator $\oplus$ to add properties to a node or messages to an inbox. We use the binding operator $\otimes$ to bind a message to the edge type it passes through.

\textbf{Algorithm.} A \ac{GraphTM} is built upon the Convolutional \ac{CoTM}~\citep{granmo2019convolutional, glimsdal2021coalesced}, with the following hyperparemeters: depth $D$, number of clauses $m$, specificity $s$, and voting margin $T$. The \ac{GraphTM} evaluates clauses and nodes in parallel as follows when given an input graph $G$: 
\begin{enumerate}
\item Evaluate each layer-zero clause component for every node in the graph: $C_j^0(v_q), j \in \{1, 2, \ldots, m\}, q \in V$.
\item \textbf{If} $C_j^0(v_q)$ \textbf{then} submit message $\boldsymbol{M}_j^0 \otimes \boldsymbol{t}$ to the neighbors of $v_q$ according to $E$, bound to the edge type $\boldsymbol{t}$.
\item \textbf{For} $i \in (1, 2, \ldots, D-1)$:
    \begin{enumerate}
        \item Evaluate each layer-$i$ clause component for every node in the graph: $C_j^i(v_q), j \in \{1, 2, \ldots, m\}, q \in V$.
        \item \textbf{If} $C_j^i(v_q)$ \textbf{then} submit message $\boldsymbol{M}_j^i \otimes \boldsymbol{t}$ to the neighbors of $v_q$ according to $E$, bound to the edge type $\boldsymbol{t}$.
    \end{enumerate}
    \item Calculate truth values of full clauses $C_j, j \in \{1, 2, \ldots, m\}$ from the clause components: $C_j = C_j^0 \land C_j^1 \land \cdots \land C_j^D$.
    \item Perform standard Coalesced \ac{TM} clause voting and update steps (See Appendix~\ref{app:CoTM} for a brief formal description of the updating scheme and Section~\ref{sec:seq} for a concrete example) using the full clauses and the complete set of properties and messages across the layers.
\end{enumerate}

\section{Experimental Results}
The selection of experiments aimed to highlight different aspects of the \ac{GraphTM}, focusing on image classification, natural language processing, and DNA sequence classification. All experiments except one\footnote{The experiments in Section \ref{KunalScalability} were conducted using a computing system with 48 CPU cores, 1460.11 GB of RAM, and a NVIDIA Tesla V100-SXM3 GPU (32 GB memory), running on a 64-bit Linux architecture.}, were executed on an NVIDIA DGX H100 system with 8 H100 GPUs (640 GB total GPU memory), 112 CPU cores, and 2 TB of RAM, running on a 64-bit Linux architecture. 

\textbf{GPU Implementation Details}. GraphTM's efficiency on CUDA GPUs stems from offloading core TM computations to custom CUDA kernels. These kernels, written in low-level code, enable highly parallel execution of Tsetlin Automata state updates, clause evaluation, and message passing directly within GPU memory. A Python interface manages data transfer and kernel execution with required configurations, critically enhancing performance for the inherently parallel nature of TM operations, 
though its reliance on custom kernels may introduce complexity compared to more generalized deep learning frameworks.



\subsection{Sequence Classification}
\label{sec:seq}
We first conducted a sequence classification experiment as a toy example to illustrate how GraphTM leverages its graph input and layered structure to capture patterns efficiently and effectively (for a thorough instance-based description of GraphTM, see App.~\ref{app:GraphTM}).
All input sequences consist of five letters, and the task is to recognize sequences containing three consecutive “A”s. The dataset includes 13,330 positive samples and 26,670 negative samples. The GraphTM consists of two layers and uses a total of four clauses to learn sub-patterns in the data, i.e., $C_j=C^0_j \wedge C^1_j,~j=\{0,1, 2, 3\}$.

After four epochs of training, the training accuracy (with $1\%$ noise) is $99.03\%$, and the test accuracy (noise free) is $100.00\%$. The training results are the sub-patterns represented in the following clauses:
\begin{tiny}
\begin{align}
&C_0= \neg A \wedge r1:0 \wedge r1:1; [3,-3], \nonumber\\
&C_1= l1:0 \wedge l1:1 \wedge l1:3 \wedge \neg r1:0; [3,-2], \nonumber\\
&C_2= A \wedge r1:2 \wedge r1:3 \wedge \neg r1:0 \wedge \neg l1:0; [-5, 6], \nonumber\\
&C_3= A \wedge l1:2 \wedge l1:3 \wedge \neg r1:0 \wedge \neg l1:0 \wedge \neg r1:1 \wedge \neg r1:2; [-2, 2], \nonumber
\end{align}
\end{tiny}

where $A$ denotes the node property at Layer 0, and notations such as $r{\color{green}1}:{\color{orange}0}$ represent messages. The number ${\color{green}1}$ before the colon denotes the \textcolor{green}{layer index}, while the number ${\color{orange}0}$ after the colon denotes the \textcolor{orange}{clause index}. The symbols $l$ and $r$ represent left and right edges, respectively. Thus, $r{\color{green}1}:{\color{orange}0}$ indicates that a message is sent through the right edge, reporting that the left neighbor matches the sub-pattern represented by Clause 0 in the previous layer. The vector at the end of each clause indicates how much the sub-pattern represented by that clause contributes to each class. The first element corresponds to the weight for class 0 (negative class), and the second corresponds to the weight for class 1 (positive class). In a multi-class setting, this vector contains multiple elements. These weights are used in the final classification stage to aggregate the contributions from all clauses that evaluate to True, and the class with the largest weighted sum is selected as the final prediction.


\begin{table}[htbp]
\centering
\caption{Clause components and clause traceability in a 2-layer \\ \ac{GraphTM} in the sequence classification experiment.}
\begin{tiny}
\begin{tabular}{r| r r| r r| r }
\toprule
$C_j$ & $C_j^0$ & $C_j^1$ & $C_j^0(X_n)$ & $C_j^1(X_n)$  & \textbf{Weights} \\
\midrule
$C_0$ & $\neg$ A & r1:0 & $C_0^0(X_n)$ & $C_0^0(X_{n-1})$ & $[3, -3]$ \\
                       &                            & r1:1 &              & $C_1^0(X_{n-1})$ & \\
\midrule
$C_1$ & $\phi$ & l1:0 & $C_1^0(X_n)$ & $C_0^0(X_{n+1})$ & $[3, -2]$\\
                       &                         & l1:1 &              & $C_1^0(X_{n+1})$ &\\
                       &                         & l1:3 &              & $C_3^0(X_{n+1})$ &\\
                       &                         & $\neg$ r1:0 &     & $\neg C_0^0(X_{n-1})$ & \\
\midrule
$C_2$ & A & r1:2 & $C_2^0(X_n)$ & $C_2^0(X_{n-1})$ & $[-5, 6]$\\
                       &                    & r1:3 &               & $C_3^0(X_{n-1})$ &\\
                       &                    & $\neg$ r1:0 &       & $\neg C_0^0(X_{n-1})$ &\\
                       &                    & $\neg$ l1:0 &       & $\neg C_0^0(X_{n+1})$ &\\
\midrule
$C_3$ & A & l1:2 & $C_3^0(X_n)$ & $C_2^0(X_{n+1})$ & $[-2, 2]$\\
                       &                    & l1:3 &               & $C_3^0(X_{n+1})$ &\\
                       &                    & $\neg$ r1:0 &       & $\neg C_0^0(X_{n-1})$ &\\
                       &                    & $\neg$ l1:0 &       & $\neg C_0^0(X_{n+1})$ &\\
                       &                    & $\neg$ r1:1 &       & $\neg C_1^0(X_{n-1})$ &\\
                       &                    & $\neg$ r1:2 &       & $\neg C_2^0(X_{n-1})$ &\\
\bottomrule
\end{tabular}
\end{tiny}
\label{table:clauseXn}
\end{table}

Table. \ref{table:clauseXn} shows the clauses in a structured manner (columns 2 and 3). The table also decoded the message captured by $C^1_j$, traced it back to the previous node layer (column 5). When a sequence is encoded as a graph and entered into \ac{GraphTM}, each node is evaluated by every clause in \ac{GraphTM}, denoted as $C_j(X_n)$. The evaluation of node $X_n$ by the component of the clause in layer $i$ is written as $C_j^i(X_n)$ (columns 4 and 5 ).

\newpage

Taking $C_0$ as an example, the evaluation at node $X_n$ is:
\begin{tiny}
\begin{align}
\label{eq:C_0}
C_0(X_n) 
&= C_0^0(X_n) \land C_0^1(X_n) \tag{a} \\
&= C_0^0(X_n) \land C_0^0(X_{n-1}) \land C_0^1(X_{n-1}) \tag{b} \\
&= \mathcal{M}(C_0^0,X_n) \land \mathcal{M}(C_0^0,X_{n-1}) \land \mathcal{M}(C_0^1,X_{n-1}) \tag{c} \\
&= \mathcal{M}(\neg A,X_n) \land \mathcal{M}(\neg A,X_{n-1}) \land \mathcal{M}(\phi,X_{n-1}) \tag{d} \\
&= \mathcal{M}(\neg A,X_n) \land \mathcal{M}(\neg A,X_{n-1}), \tag{e}
\end{align}
\end{tiny}

where (a) follows from the definition of $C_0=C_0^0 \wedge C_0^1$.
(b) is obtained by tracing the message back to the previous layer. $C^1_0=r1:0 \wedge r1:1$, where ${\color{blue}r}{\color{green}1}:{\color{orange}0}$ means that a message from the {\color{blue}right} edge tells $X_n$ that the {\color{blue}left} neighbor {\color{blue}$X_{n-1}$} matches clause component $C_{\color{orange}0}^{\color{green}0}$ at the previous layer (Layer {\color{green}$0$}), hence ${\color{blue}r}{\color{green}1}:{\color{orange}0} = C_{\color{orange}0}^{\color{green}0}({\color{blue}X_{n-1}})$. Similarly, 
${\color{blue}r}{\color{green}1}:{\color{orange}1} = C_{\color{orange}1}^{\color{green}0}({\color{blue}X_{n-1}})$.
(c) introduces the matching operator $\mathcal{M}$\footnote{should not be confused with the message denotation $M$.}, representing the matching operation between its two arguments. If $X_n$ is a boundary node, either $X_{n-1}$ or $X_{n+1}$ does not exist. The matching result with a non-existent node is always False, except when matching with an empty clause $\phi$.
(d) substitutes the clause definitions, and (e) eliminates the empty clause $\phi$.


Tracing messages back along edges applies to all message layers: regardless of depth, any message can be traced layer by layer to the node layer (see Table~\ref{table:exp2clause}). This enables message-layer features to be expressed as intuitive node-layer properties, which makes it possible to explain how GraphTM performs feature matching for classification and is key to GraphTM’s interpretability.

The four clause evaluations at node $X_n$ are:
\begin{tiny}
\begin{align}
\label{eq:C_0-3}
C_0(X_n) 
= & \mathcal{M}(\neg A,X_n) \land \mathcal{M}(\neg A,X_{n-1}),\\
 C_1(X_n) 
= & \mathcal{M}(\neg A, X_{n+1}) \wedge \mathcal{M}(A, X_{n+1}) \wedge \neg \mathcal{M}(\neg A, X_{n-1}), \nonumber\\    
 C_2(X_n) 
= &\mathcal{M}(A, X_n) \wedge \mathcal{M}(A, X_{n-1}) \wedge \neg \mathcal{M}(\neg A, X_{n-1}) \nonumber\\
&\wedge \mathcal{M}(\neg A, X_{n+1}), \nonumber\\
C_3(X_n) 
= & \mathcal{M}(A, X_n) \wedge \mathcal{M}(A, X_{n+1}) \wedge \neg \mathcal{M}(\neg A, X_{n-1})  \nonumber\\
& \wedge \neg \mathcal{M}(\neg A, X_{n+1}) \wedge \neg \mathcal{M}(\phi, X_{n-1}) \wedge \neg \mathcal{M}(A, X_{n-1}). \nonumber
\end{align}
\end{tiny}

Assuming that a test sequence ``BAAAE'' enters to the \ac{GraphTM}, then by applying Eqs. \ref{eq:C_0-3}, we obtain Table \ref{table:clause_on_node}, displaying the evaluation results of each clause on each node. Taking $C_1(X_0)$ as an example to illustrate how the result is derived: 
\begin{tiny}
\begin{align}
    & C_1(X_0) \nonumber\\
    = & \mathcal{M}(\neg A, X_{1}) \wedge \mathcal{M}(A, X_{1}) \wedge \neg \mathcal{M}(\neg A, X_{-1}) \nonumber\\
    = & \mathcal{M}(\neg A, A) \wedge \mathcal{M}(A, A) \wedge \neg \mathcal{M}(\neg A, X_{-1}) \nonumber\\
    = & False \wedge True \wedge \neg False \nonumber\\
    = & False \nonumber
\end{align}
\end{tiny}

\begin{table}[htbp]
\centering
\begin{tiny}
\caption{Clause evaluation results at each node, when the input sequence is ``BAAAE''.}
\begin{tabular}{cccccc}
\toprule
$C_j(X_n)$ & $X_0$=B & $X_1$=A & $X_2$=A & $X_3$=A & $X_4$=E \\
\midrule
$C_0$ & $False$ & $False$ & $False$ & $False$ & $False$ \\
$C_1$ & $False$ & $False$ & $False$ & $False$ & $False$ \\
$C_2$ & $False$ & $False$ & $False$ & $True$  & $False$ \\
$C_3$ & $False$ & $False$ & $False$ & $False$ & $False$ \\
\bottomrule
\end{tabular}
\label{table:clause_on_node}
\end{tiny}
\end{table}

$C_j(X_n)$ denotes the evaluation of clause $j$ on a single node. The evaluation on the entire graph is therefore $M_j=C_j(X_0) \vee C_j(X_1) \vee C_j(X_2) \vee C_j(X_3) \vee C_j(X_4)$, meaning that if any node matches the clause, the entire graph is considered to match it, and the clause evaluates to $True$ on the graph. This is straightforward to interpret: each node receives messages from its neighbors, and when a clause evaluates to $True$, it indicates not only that the node itself has the required property, but also that its neighbors do as well, corresponding here to three consecutive `A's. Based on Table \ref{table:clause_on_node}, we have: $M_0=False$, $M_1=False$, $M_2=True$, and $M_3=False$. 

We can now apply the weighted sum approach to compute the total weight of all clauses that evaluate to $True$, which in this example is $[-5,6]$. The class corresponding to the largest total weight is the final predicted class, i.e., class~1, indicating the input graph is a positive sample. The \ac{GraphTM} has thus correctly predicted that the sequence ``BAAAE'' contains three consecutive ``A''.



\subsection{Disconnected Nodes}
\begin{figure}[htbp]
    \centering
    \includegraphics[width=0.9\linewidth]{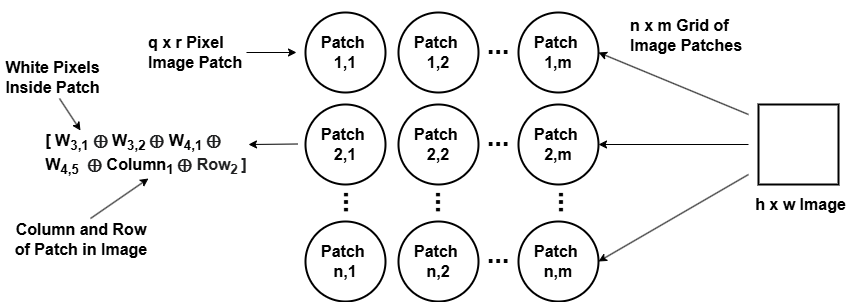}
    \caption{Encoding images into graphs.}
    \label{fig:mnistconv_input}
\end{figure}

The goal of these experiments is to verify that the \ac{GraphTM} effectively reduces to the convolutional variant of \ac{CoTM}~\cite{glimsdal2021coalesced} in scenarios where the nodes are disconnected with no connections between them. In order of increasing complexity: MNIST~\cite{lecun1998gradient}, Fashion MNIST (F-MNIST)~\cite{xiao2017fashion} and CIFAR-10~\cite{krizhevsky2009learning} were examined here. For MNIST, images were initially binarized using a stationary threshold before being converted to disconnected graphs. For F-MNIST, binarization was done using thermometer encoding with eight bins~\cite{gronningsaeter2024toolbox}. Finally, for CIFAR-10, both adaptive Gaussian thresholding and color thermometer encoding using eight bins were used to convert images to binary representation. The CIFAR-10 dataset was further expanded by applying horizontal flipping. To encode the datasets into the disconnected graph structure shown in Figure~\ref{fig:mnistconv_input}, each image was split into patches. These patches and their position were then encoded into a node.

The \ac{GraphTM} and \ac{CoTM} were trained using identical hyperparameters for 30 epochs, with the average accuracy over the final five epochs presented in Table~\ref{tab:result-mnist}. Due to the absence of edges connecting the nodes in the graph encoding, we would expect the \ac{GraphTM} model to exhibit similar behavior to the \ac{CoTM}. This expectation is confirmed by the comparable results observed for MNIST and F-MNIST. However, the \ac{GraphTM} outperforms the \ac{CoTM} on CIFAR-10. This improvement is likely due to the \ac{GraphTM}’s ability to incorporate multiple views of each image. While the \ac{GraphTM} could also have been trained using only adaptive Gaussian thresholding, multiple complementary encodings were intentionally employed to highlight the advantages of the graph-based input. In contrast, the \ac{CoTM} processes each image as a single feature tensor, requiring all information to be encoded within the same representation. This prevents the direct use of multiple heterogeneous encodings without merging them at the input level. As a result, the \ac{CoTM} was trained using only adaptive Gaussian thresholding.



\begin{table}[htbp]
\centering
\caption{Classification accuracies (\%) for the MNIST, F-MNIST, and CIFAR-10 datasets.}
\label{tab:result-mnist}
 \begin{small}
\begin{tabular}{ l c c c }
\toprule
    Model      & MNIST  & F-MNIST & CIFAR-10 \\ 
\midrule
\ac{GraphTM}  & 98.42 $\pm$ 0.05  & 89.49 $\pm$ 0.09  & \textbf{70.28 $\pm$ 0.17} \\ 
\ac{CoTM}  & \textbf{98.93 $\pm$ 0.02}  & \textbf{91.05 $\pm$ 0.09}   & 66.42 $\pm$ 0.19   \\ 
\bottomrule
\end{tabular}
\end{small}
\end{table}

\begin{figure}[htbp]
    \centering
    \includegraphics[width=0.95\linewidth]{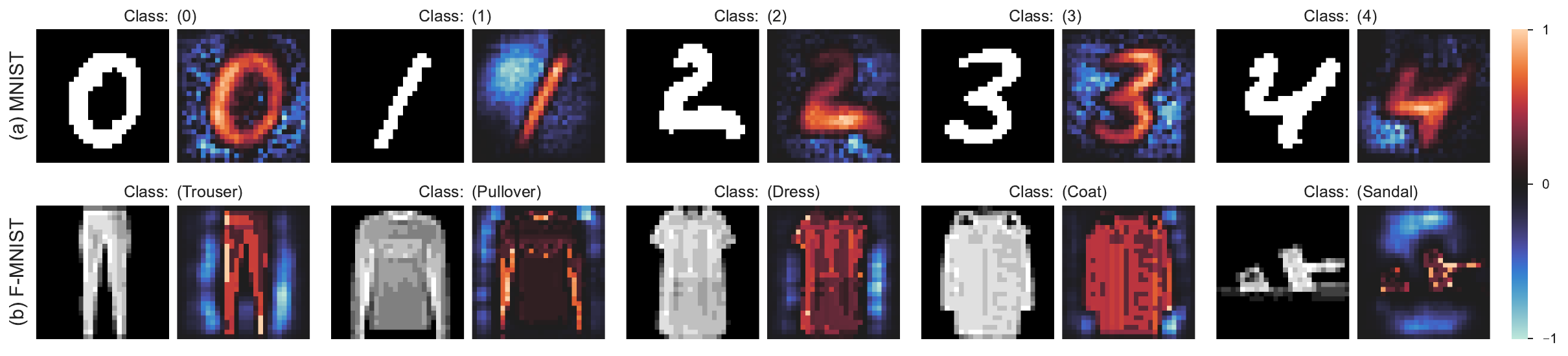}
    \caption{Interpreting the active clauses for a given input from the (a) MNIST and (b) F-MNIST datasets. The red region shows the activations for the symbols representing white pixels. Whereas the blue region represent the absence/black pixels.}
    \label{fig:conv_clauses}
\end{figure}

We also reviewed the interpretability of the \ac{GraphTM}. The clauses learned by the \ac{GraphTM} are in a hypervector format. Consequently, the symbols encoded in the graph can be \textit{ANDed} with the clause, to check if the symbol is included in the clause. 
Since the symbols represent the pixels in an image patch and its location, the symbols in the active clauses can be placed in their appropriate positions. Figure~\ref{fig:conv_clauses} illustrate the learned patterns, by aggregating the active clauses for a given input image.



\subsection{Connected Nodes with Superpixels}
\label{subsec_superpixels}

We evaluated \ac{GraphTM} on the MNIST Superpixel dataset~\cite{monti2017geometric} and compared it to two \ac{GCN} models. The images were represented as graphs with superpixels as nodes and the edges being defined by spatial adjacency obtained via k-NN. Node features included X and Y centroid coordinates and average grayscale intensity of the grouped pixels. We quantized grayscale intensities to eight linear levels. Furthermore, adding the X and Y coordinates of the nearest neighbor and a count of neighbors as features to each node were shown to enhance the accuracy with about 3\% and 5\%, respectively. Demonstrated in Table~\ref{tab:result-MNIST-superpx}, the \ac{GraphTM} achieved an accuracy of 89.24 $\pm$ 1.34~\%, 
compareable with \ac{GCN} from 2016~\cite{monti2017geometric}, reporting 75.62\% with ChebNet~\cite{defferrard2016chebnet}, and 91.11\% with MoNet.

\begin{table}[htbp]
\centering
\caption{Classification accuracies (\%) for the MNIST Superpixel dataset.}
\label{tab:result-MNIST-superpx}
\begin{small}
\begin{tabular}{ l c c c c }
\toprule
Model    & \ac{GraphTM} (2000 clauses) & ChebNet & MoNet \\ 
\midrule
Acc. (\%) & 88.09 $\pm$ 0.27                      & 75.62       & \textbf{91.11}     \\ 
\bottomrule
\end{tabular}
\end{small}
\end{table}

\subsection{Sentiment Polarity Classification}
\label{subsec:sentiment}

This experiment deals with benchmarking the performance of the \ac{GraphTM} with respect to three well-known datasets for determination of sentiment polarity in Natural Language Processing. The datasets used in this case were IMDB~\cite{maas2011learning}, Yelp~\cite{zhang2015character-level} and MPQA~\cite{wiebe2005annotating}. Each dataset was converted into a binary classification problem when necessary. 

\begin{figure}[htbp]
    \centering
    \includegraphics[width=0.7\linewidth]{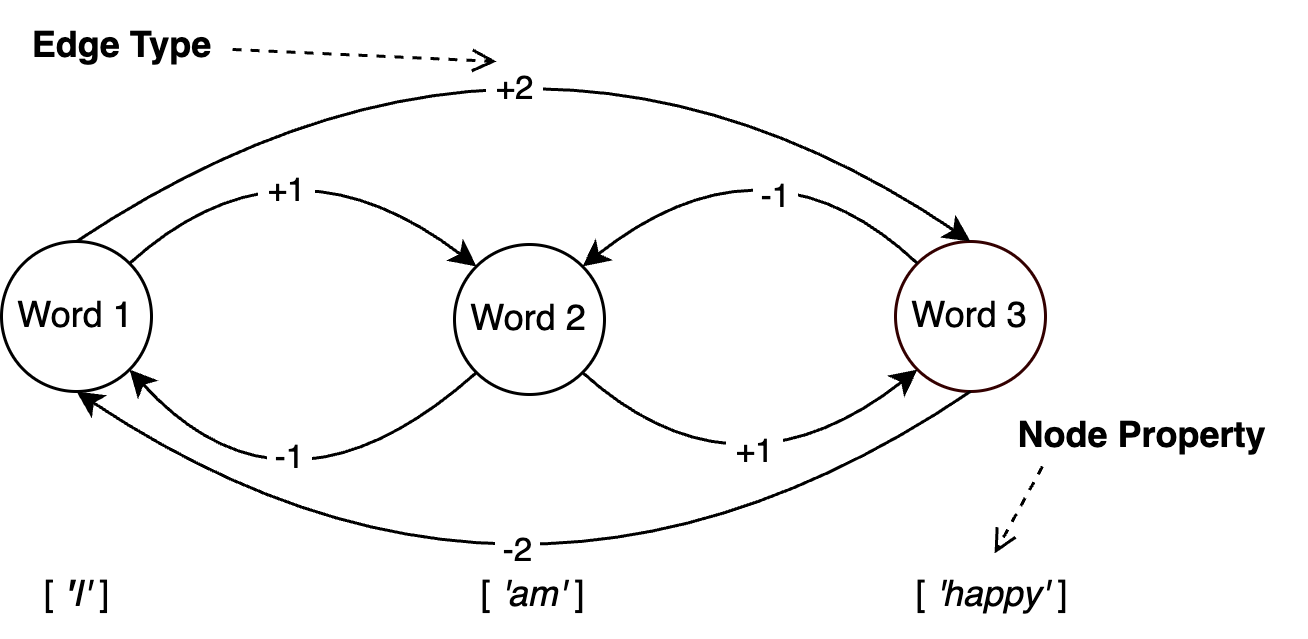}
    \caption{Graph structure representation for sentences.}
    \label{fig:Graph structure}
    
\end{figure}

\begin{table*}[htbp]
\centering
\caption{Accuracy obtained by \ac{GraphTM} with depth 1 and depth 2, vs that obtained by \ac{GNN} and Standard TM (StdTM) for three different datasets for sentiment polarity.}
\label{tab:sentimentpolarity}
\begin{small}
\begin{tabular}{ l c c c c }
\toprule
Datasets & \ac{GraphTM} (Depth = 1) & \ac{GraphTM} (Depth = 2) & \ac{GNN} & StdTM \\ 
\midrule
Imdb              & 86.43 $\pm$ 2.10            & 88.15 $\pm$ 2.16          & \textbf{89.19} $\pm$  2.08     & 84.60 $\pm$  0.55            \\ 
Yelp              & 84.15 $\pm$ 1.32             & 85.24  $\pm$ 1.45            & 84.16 $\pm$ 0.92      & \textbf{86.69}  $\pm$ 0.50            \\ 
MPQA              & 80.92  $\pm$ 2.30            & 81.77  $\pm$ 1.15            & \textbf{83.73} $\pm$ 0.47       & 70.06  $\pm$ 0.24             \\ 
\bottomrule
\end{tabular}
\end{small}
\end{table*}

For each dataset, the same graph encoding method was followed, where one word and its position was represented as a node, and each node was connected to every other node in a fully connected graph structure as depicted in Figure~\ref{fig:Graph structure}. Further, the edge-type between nodes was their word distance between each other in the sentence, with left and right indicated as the polarity of that distance (e.g., two words to the right is +2, two words to the left is -2). For the TM, words are encode without positional encoding. 

We compared the performance of \ac{GraphTM} with a \ac{GNN} and a standard \ac{TM}. Across all three datasets, \ac{GraphTM} with depth 2 outperformed depth 1, indicating that nested clauses capture more relevant information. \ac{GraphTM} with depth 2 achieved accuracy close to that of the \ac{GNN}. Compared to the standard \ac{TM}, \ac{GraphTM} achieved higher accuracy on IMDB and MPQA, while the standard TM performed better on Yelp.

%



\subsection{Tracking Action Coreference} 
\label{subsec:actioncoreference}
\begin{figure}[htbp]
    \centering
\includegraphics[width=0.5\linewidth]{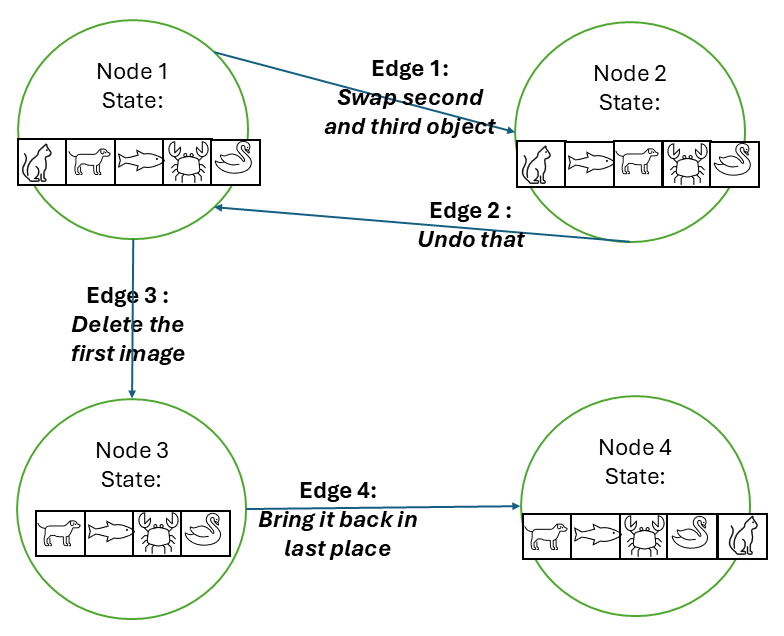}
    \caption{Graph construction for action coreference tracking.}
    \label{fig:action coref}
\end{figure}

In this experiment we used a modified version of the Tangram dataset from the Scone dataset~\cite{long2016simpler}, with the goal of tackling the linguistic phenomenon of action coreference. Each example consists of a sequence of 5 items and a sequence of 5 actions performed on the items. The actions are one of the following: swap, delete, or bring back. Each action is expressed in natural language terms, and is also referred to as an ``utterance''. A typical setup, as illustrated in Figure~\ref{fig:action coref}, can be as follows: ``Given an ordered set of five different images. Swap the second and third object. Undo that. Delete the first image. Bring it back in last place.''

The \ac{GraphTM} was tasked with determining which one of the images occupied position $x$, after the sequence of actions had been performed. The challenge lies in the fact that even though there are only 3 unique actions, there are over 9,454 different textual instructions corresponding to them. For example, ``delete'' can be referred to as ``remove 1st object'', or ``take away the second image'', or ``get rid of the item in the middle''. As graph input to the \ac{GraphTM}, each action were represented as a connection between the graph nodes, while graph nodes themselves were the states (object sequences) before and after the action was performed. 

The results are compared with author-reported values, without independent verification~\cite{guu2017language}. The competing methods are broadly based on Reinforcement Learning (REINFORCE), maximum marginal likelihood (BS-MML) and randomized beam search (RANDOMER). In order to compare the performance of \ac{GraphTM} with these methods, examples of length 3 utterance and 5 utterances were used as input while the final resultant sequence was the expected output of those. The obtained accuracy values are shown in Table~\ref{tab:result-actioncorref}.
This experiment explores two aspects: (1) using the \ac{GraphTM} with input graphs characterized by few unique nodes and a large number of unique edges, and (2) evaluating \ac{GraphTM} in sequence understanding for text. For experiments with 3 utterance length, the results showed by the \ac{GraphTM} are slightly better than BS-MML, but do not manage to exceed either the Randomized Beam Search nor the Reinforcement Learning based model. However, the \ac{GraphTM} outperforms the other methods in case of experiments with 5 utterance lengths. 

\begin{table}[htbp]
\centering
\caption{Accuracy on 3 and 5 utterance lengths obtained via \ac{GraphTM} vs other reported methods~\cite{guu2017language}.}
\label{tab:result-actioncorref}
\begin{small}
\begin{tabular}{ l c c }
\toprule
   Model                & 3 utterance & 5 utterance \\ 
\midrule
GraphTM      & 64.14 $\pm$ 1.69        & \textbf{57.92 $\pm$ 1.02}         \\ 
REINFORCE & \textbf{68.5}          & 37.3          \\ 
BS-MML    & 62.6          & 32.2          \\ 
RANDOMER  & 65.8          & 37.1          \\ 
\bottomrule
\end{tabular}
\end{small}
\end{table}

\subsection{Recommendation Systems}

\begin{figure}[htbp]
    \centering
    \includegraphics[width=0.6\linewidth]{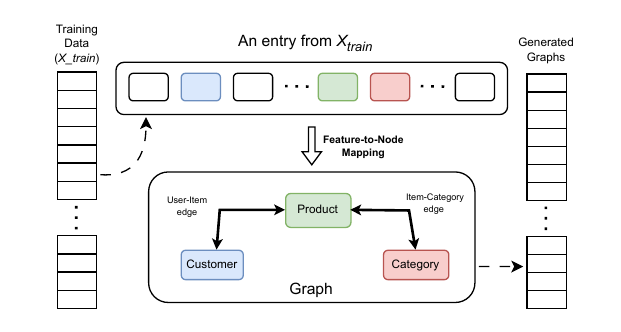}
    \caption{Graph construction for recommendation systems: representing customer, product, category relations from training data.}
    \label{fig:recomm_sys}
\end{figure}

This study investigates the application of \ac{GraphTM} to extract latent relationships between features in databases for recommendation systems. 
In this experiment, the emphasis was placed on modeling three interconnected entities within the dataset: customers, products, and product categories. The \ac{GraphTM} model was trained using labeled data in the form of user-provided rankings or scores, which represented the preferences or satisfaction levels associated with specific customer-product-category combinations (see Figure~\ref{fig:recomm_sys}). 

The experiment employed three machine learning models: Standard TM (StdTM), \ac{GraphTM}, and \ac{GCN}. All experiments were conducted on a publicly available database 
obtained via Amazon~\cite{kaggleamazon}. It was expanded tenfold, with varying levels of added noise introduced to evaluate its impact on accuracy. Table~\ref{tab:recomm_sys_accuracy} presents the accuracy achieved by the three models with different noise ratios.
Overall, \ac{GraphTM} demonstrates competitive performance across all noise levels, closely matching the accuracy of \ac{GCN} while substantially outperforming the standard \ac{TM}. Notably, although \ac{GCN} attains the highest accuracy in most settings, the performance gap between \ac{GCN} and \ac{GraphTM} remains small even at higher noise levels, indicating that \ac{GraphTM} maintains robust predictive capability in noisy environments.
\ac{GraphTM} also exhibits superior computational efficiency, particularly in a CUDA-enabled environment, achieving competitive results with a total time of 133.75 seconds compared to standard \ac{TM}'s 1,068.99 seconds. 

\begin{table}[htbp]
\centering
\caption{Accuracy of \ac{GCN}, \ac{GraphTM}, and Standard \ac{TM} (StdTM) under varying noise ratios.}
\label{tab:recomm_sys_accuracy}
\begin{small}
\begin{tabular}{l c c c c c c}
\toprule
 & \multicolumn{6}{c}{Noise Ratio} \\
\cmidrule(lr){2-7}
Model
 & 0.005 & 0.01 & 0.02 & 0.05 & 0.1 & 0.2 \\
\midrule

\ac{GCN} & \shortstack{ 99.52 \\ \scriptsize $\pm$0.40 } & \shortstack{ 98.96 \\ \scriptsize $\pm$0.28 } & \shortstack{ 98.36 \\ \scriptsize $\pm$0.27 } & \shortstack{ 95.80 \\ \scriptsize $\pm$0.25 } & \shortstack{ 91.11 \\ \scriptsize $\pm$0.27 } & \shortstack{ 83.45 \\ \scriptsize $\pm$1.05 } \\
\ac{GraphTM} & \shortstack{ 98.73 \\ \scriptsize $\pm$0.12 } & \shortstack{ 98.35 \\ \scriptsize $\pm$0.08 } & \shortstack{ 97.73 \\ \scriptsize $\pm$0.13 } & \shortstack{ 94.61 \\ \scriptsize $\pm$0.34 } & \shortstack{ 89.85 \\ \scriptsize $\pm$0.29 } & \shortstack{ 78.73 \\ \scriptsize $\pm$0.75 } \\
StdTM & \shortstack{ 76.73 \\ \scriptsize $\pm$0.14 } & \shortstack{ 74.87 \\ \scriptsize $\pm$0.12 } & \shortstack{ 72.24 \\ \scriptsize $\pm$0.26 } & \shortstack{ 63.86 \\ \scriptsize $\pm$0.34 } & \shortstack{ 49.48 \\ \scriptsize $\pm$0.38 } & \shortstack{ 20.13 \\ \scriptsize $\pm$0.04 } \\

\bottomrule
\end{tabular}
\end{small}
\end{table}


Another RS experiment was run on the MovieLens dataset, formulating it as a top-n recommendation problem. GraphTM outperforms a set of well-established dedicated RS algorithms. Please refer to Appendix~\ref{app:topn} for details.

\subsection{Viral Genome Sequence Data} \label{KunalScalability}

We present the performance of the \ac{GraphTM} on viral disease classification using nucleotide sequences and compare it with \acp{NN}. Computational efficiency and scalability are also examined. The dataset consists of labeled sequences from a publicly available nucleotide sequence database~\cite{Sequence_d}. 
Pre-processing involved removing nonstandard nucleotide samples, therefore leaving samples with only: A, C, G, and T. 


Initial accuracy tests were conducted on a balanced dataset comprising 8,995 samples (1,799 per class) from five virus classes: Influenza A virus, SARS-CoV-2, Dengue virus, Zika virus, and Rotavirus. Since only 1,799 samples were available for SARS-CoV-2, the same number was used for all classes. Genome sequences were truncated to the first 500 nucleotides and encoded using 3-mers to capture nucleotide patterns. BiLSTM, LSTM, BiLSTM-CNN, GRU, and \ac{GCN} were used as comparable \ac{NN} models.

Table~\ref{tab:performance_comparison} presents the classification performance of different methods on the 5-class dataset after 10 epochs. Among the \ac{GraphTM} variants, \ac{GraphTM} with depth 2 significantly outperformed GraphTM-1, achieving 95.17\% training accuracy and 95.14\% testing accuracy with a training time of 84.37 seconds. The GCN model achieved the highest overall testing accuracy (96.35\%) and a training accuracy of 96.64\%; however, it required substantially more training time (226.36 seconds). Notably, the GCN was trained in a sample-by-sample manner, similar to the GraphTM, whereas the other neural network models were trained using mini-batch optimization. BiLSTM-CNN also demonstrated strong performance, reaching 96.77\% training accuracy and 95.44\% testing accuracy with a comparatively shorter training time of 32.65 seconds. Simpler recurrent models such as LSTM and GRU showed lower classification accuracy but benefited from significantly reduced training times, with GRU slightly outperforming LSTM. Overall, the \ac{GraphTM} method, especially with depth 2, highlights the competitiveness of the method, providing a good balance between accuracy and training time. 

\begin{table*}[htbp]
  \caption{Classification accuracies and training times on the dataset with 5 classes, after 10 epochs.}
  \label{tab:performance_comparison}
  \centering
  \begin{small}
  \begin{tabular}{lccccccc}
    \toprule
    Metric & GraphTM-1 & GraphTM-2 & BiLSTM & LSTM & GRU & BiLSTM-CNN & GCN \\
    \midrule
    Training Accuracy (\%) & 60.74 & 95.17 & 94.43 & 88.70 & 94.68 & \textbf{96.77} & 96.64 \\
    Testing Accuracy (\%)  & 59.81 & 95.14 & 92.69 & 87.29 & 94.05 & 95.44 & \textbf{96.35} \\
    Training Time (s)      & 62.47 & 84.37 & 50.39 & 26.02 & \textbf{25.47} & 32.65 & 226.36 \\
    \bottomrule
  \end{tabular}
  \end{small}
\end{table*}

The \ac{GraphTM}'s performance when faced with increasing class complexity was also studied with class ranges from two to five classes, involving combinations of Influenza A virus, SARS-CoV-2, Dengue virus, Zika virus, and Rotavirus. As shown in Table~\ref{table:clause_effect}, as the number of clauses increased, classification accuracy improved across all configurations. This demonstrates that when increasing class complexities was encountered, the accuracy improved with the number of clauses.

\begin{table}[htbp]
  \caption{Scalability analysis of GraphTM with increasing clause size and class complexity levels.}
  \label{table:clause_effect}
  \centering
  \begin{small}
  \begin{tabular}{lcccc}
    \toprule
    & \multicolumn{4}{c}{Clauses}\\
    \cmidrule(lr){2-5}
    Classes & 500 & 700 & 1,000 & 2,000 \\
    \midrule
    2 & 100 & 100 & 100 & 100 \\
    3 & 95.09 & 96.80 & 96.66 & \textbf{97.31} \\
    4 & 89.16 & 91.96 & 92.64 & \textbf{94.67} \\
    5 & 90.52 & 92.72 & 93.85 & \textbf{95.14} \\
    \bottomrule
  \end{tabular}
  \end{small}
\end{table}

\begin{figure}[htbp]
  \centering
  \includegraphics[width=0.8\linewidth]{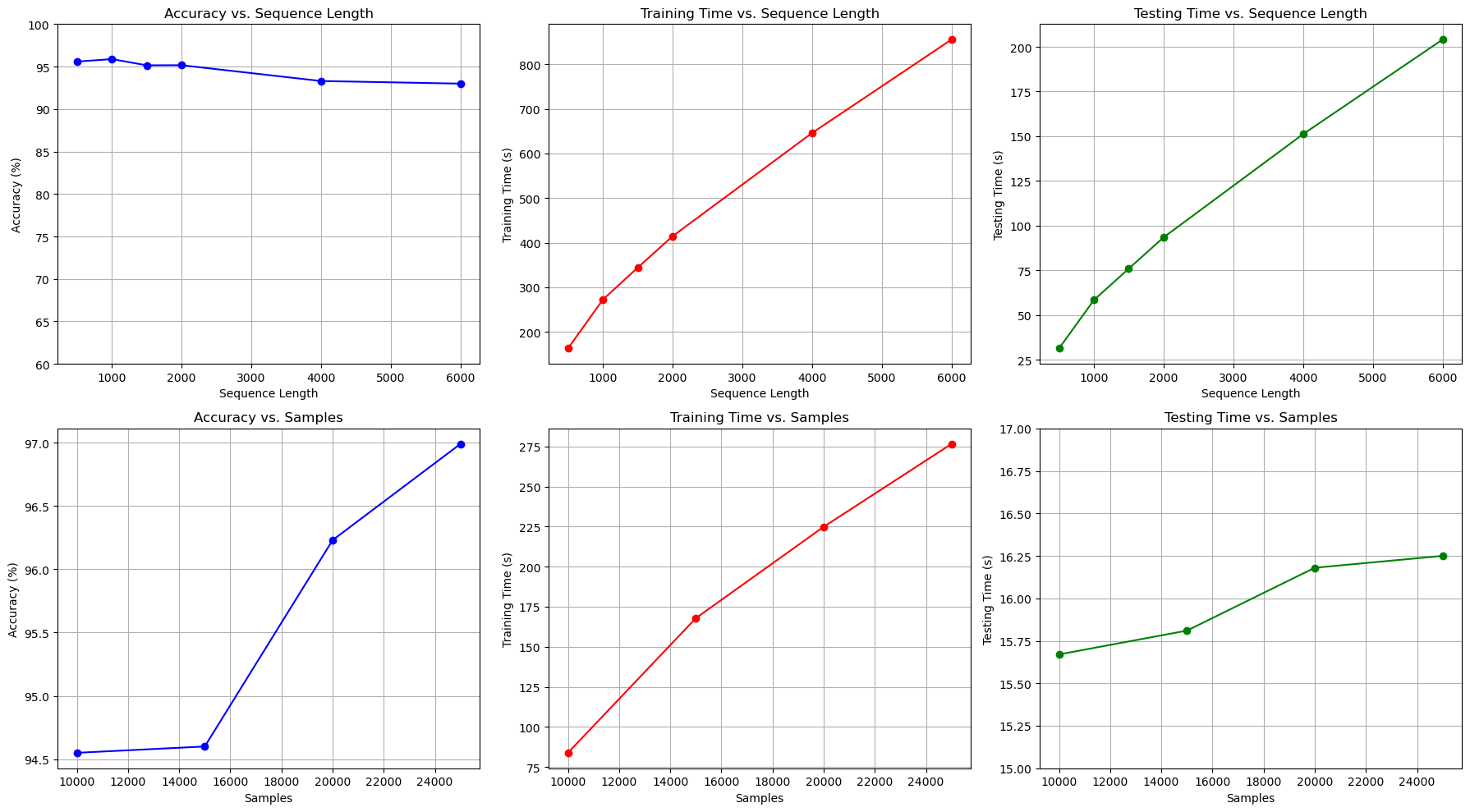}
  \caption{Scalability of GraphTM with increasing data volume and sequence length.}
  \label{fig:DNA_Sample_and_length}
\end{figure}

The computational efficiency of \ac{GraphTM} was assessed through two scalability studies. The first study evaluated training time, testing time, and test accuracy as the number of samples increased from 10,000 to 25,000, with balanced classes across five virus categories. As shown in Figure~\ref{fig:DNA_Sample_and_length}, test accuracy improved consistently, from 94.55\% at 10,000 samples to 96.99\% with 25,000 samples. The second study examined the impact of sequence length on performance, with lengths ranging from 500 to 6,000. As illustrated in Figure~\ref{fig:DNA_Sample_and_length}, test accuracy peaked at 95.88\% for a length of 1,000, then gradually declined to 92.99\% at 6,000. These findings demonstrate GraphTM’s scalability and computational efficiency under varying data conditions. 
The scalability assessment shows that \ac{GraphTM} maintains high accuracy and handles computational demands effectively, even as dataset size, sequence length, and class complexity increase.

\subsection{Multivalue Noisy XOR}\label{sec:mvxor}
This dataset is a modified version of the NoisyXOR dataset~\cite{granmo2018tsetlin}, where the two input variables are non-binary, and the output label is determined by applying an XOR operation to a defined relationship between them. In this case, the relation is defined to be divisibility by 2. Formally, the input space consists of two features, $x_1, x_2 \in \{0, 1, 2, \dots, n\}$, and the output label $y$ is defined by Eq.~\ref{eq:mvxor}. Consequently, the number of symbols in the graph becomes $n$. Outlined previously in Section \ref{sec:gtm}, Figure~\ref{fig:graph_tsetlin_machine_conceptual} shows the graph encoding for this dataset, consisting of two nodes, $x_1$ and $x_2$, connected by a bi-directional edge of the same type.

\begin{equation}
        y=\begin{cases}
        0 & x_1 + x_2 = 2k \quad \forall k \in \mathbb{N}\\
        1 & \text{otherwise.}
        \end{cases}
        \label{eq:mvxor}
\end{equation}

\begin{figure}[htbp]
    \centering
    \includegraphics[width=0.9\linewidth]{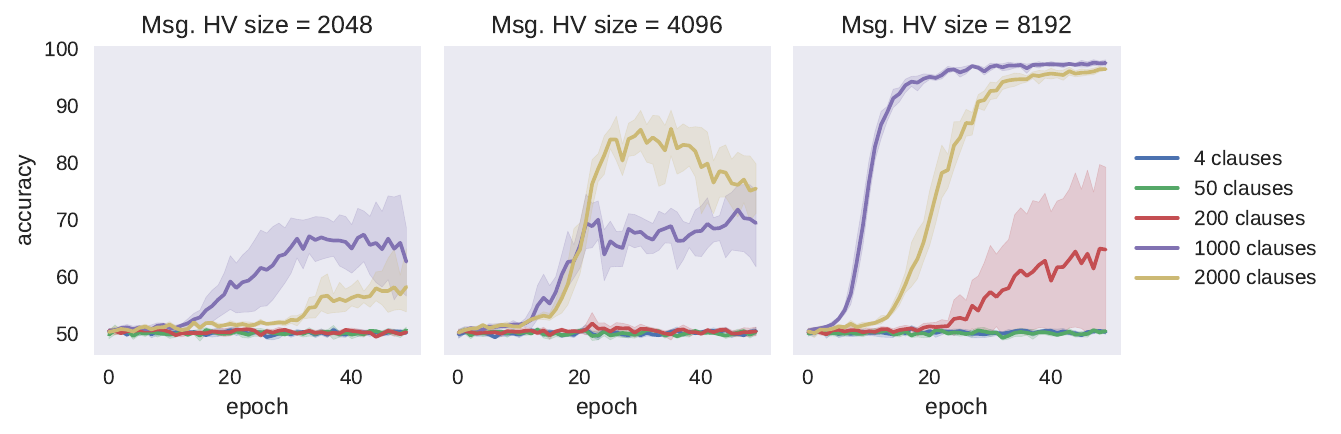}
    \caption{Test set average accuracy and standard error across five independent trials for the multivalue noisy XOR dataset with varying number of clauses and message sizes.}
    \label{fig:mvxor_hvsize}
\end{figure}

Using the Multivalue Noisy XOR dataset, we analyzed how GraphTM learning responds to varying message sizes and different number of clauses, with $n$ possible symbols available for the graph nodes.
Figure~\ref{fig:mvxor_hvsize} illustrates our findings on the influence of message size and number of clauses on learning performance. When the number of symbols is large (e.g., 500), a higher number of clauses is necessary to accurately capture all patterns. The addition of more clauses means that the message hypervector becomes more dense, and the results indicate that a larger message size is needed for enhanced learning. Interestingly, when using the largest message size, the accuracy is higher for 1,000 clauses compared to 2,000 clauses. One possible explanation is that using 2,000 clauses results in too many collisions in the message hypervectors, adversely affecting the accuracy. 

\section{Conclusions} 
The \ac{GraphTM} represents a substantial advancement of the \ac{TM} framework by enabling the learning of interpretable deep clauses from graph structured data. This capability allows \ac{GraphTM} to generalize beyond fixed-length inputs, a key limitation of existing \ac{TM} approaches. Across diverse tasks, we have demonstrated that the \ac{GraphTM} has both interpretability and strong empirical performance: it achieves a 3.86\%-points higher accuracy on CIFAR-10 compared to \ac{CoTM}s, and significantly outperforms standard \ac{TM} in sentiment polarity classification on the IMDB and MPQA datasets. In action coreference tracking tasks involving sequences of 5 utterances, it surpasses traditional reinforcement, with 19.3 percentage points in increased accuracy, underscoring its effectiveness on complex graph-based tasks. For recommendation systems, \ac{GraphTM} shows robustness to noise, comparable to a \ac{GCN}. In viral genome sequence analysis, it achieves competitive accuracy with BiLSTM-CNN and \ac{GCN}, while offering substantially faster training than \ac{GCN}. These results highlight \ac{GraphTM} as a powerful, interpretable and efficient model for various tasks, opening new avenues for advancing \ac{TM}-based approaches. 

\section*{Impact Statement}
We do not foresee significant negative societal impacts arising from this work.

\bibliography{references}
\bibliographystyle{icml2026}
\newpage
\appendix

\section{Appendix / supplemental material}

\subsection{Standard Tsetlin Machine}
\label{app:TM}

We briefly introduce here the learning entities in a Tsetlin Machine (TM), i.e., Tsetlin Automata~(TAs) and then the operational concept of the standard TM. The GraphTM will be explained in detail in \ref{app:GraphTM}. One can choose to read \ref{app:GraphTM} first.

\subsubsection{TA}
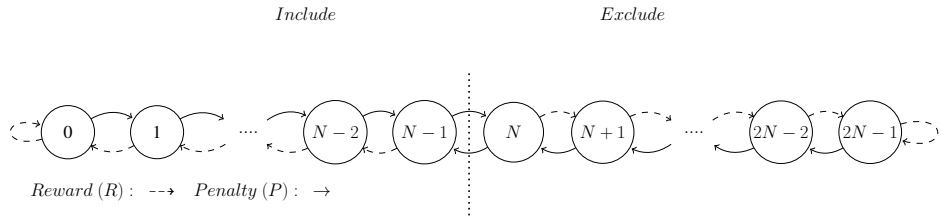
\begin{figure*}[b]
\centering
\resizebox{0.8\textwidth}{!}{
\begin{minipage}{1\textwidth}
\begin{tikzpicture}[node distance = .35cm, font=\Huge]
    \tikzstyle{every node}=[scale=0.35]
    \node[state] (A) at (0,2) {~~~~0~~~~~};
    \node[state] (B) at (1.5,2) {~~~~1~~~~~};
    
    \node[state,draw=white] (M) at (3,2) {~~~$....$~~~};
    
    \node[state] (C) at (4.5,2) {$~N-2~$};
    \node[state] (D) at (6,2) {$~N-1~$};
    
    \node[state] (E) at (7.5,2) {$~~~~N~~~~$};
    \node[state] (F) at (9,2) {$~N+1~$};
    
    \node[state,draw=white] (G) at (10.5,2) {~~~$....$~~~};
    
    \node[state] (H) at (12,2) {$2N-2$};
    \node[state] (I) at (13.5,2) {$2N-1$};

    \node[thick] at (4,4) {$Include$};
    \node[thick] at (9.5,4) {$Exclude$};
    
    \node[thick] at (1.9,1) {$Reward~(R):~\dashrightarrow$~~~~$Penalty~(P):~\rightarrow$};

    \draw[every loop]
    (A) edge[bend left] node [scale=1.2, above=0.1 of B]{} (B)
    (B) edge[bend left] node  [scale=1.2, above=0.1 of M] {} (M)
    (M) edge[bend left] node  [scale=1.2, above=0.1 of C] {} (C)
    (C) edge[bend left] node [scale=1.2, above=0.1 of D] {} (D)
    (D) edge[bend left] node  [scale=1.2, above=0.1 of E] {} (E);

    \draw[every loop]
    (I) edge[bend left] node [scale=1.2, below=0.1 of H] {} (H)
    (H) edge[bend left] node  [scale=1.2, below=0.1 of G] {} (G)
    (G) edge[bend left] node [scale=1.2, below=0.1 of F] {} (F)
    (F) edge[bend left] node  [scale=1.2, below=0.1 of E] {} (E)
    (E) edge[bend left] node  [scale=1.2, below=0.1 of D] {} (D);

    
    \draw[dashed,->]
    (B) edge[bend left] node  [scale=1.2, above=0.1 of A] {} (A)
    (M) edge[bend left] node [scale=1.2, above=0.1 of B] {} (B)
    (C) edge[bend left] node [scale=1.2, above=0.1 of M] {} (M)
    (D) edge[bend left] node [scale=1.2, above=0.1 of C] {} (C)
    (A) edge[loop left] node [scale=1.2, below=0.1 of D] {} (D);
    
    \draw[dashed,->]

    (H) edge[bend left ] node [scale=1.2, below=0.1 of I] {} (I)
    (G) edge[bend left] node  [scale=1.2, below=0.1 of H] {} (H)
    (F) edge[bend left] node  [scale=1.2, below=0.1 of G] {} (G)
    (E) edge[bend left ] node [scale=1.2, below=0.1 of F] {} (F)
    (I) edge[loop right] node [scale=1.2, below=0.1 of E] {} (E);
    
      \draw[dotted, thick] (6.75,0.6) -- (6.75,3);

\end{tikzpicture}
\end{minipage}
}
\caption{A two-action Tsetlin automaton with $2N$ states~\cite{jiao2021convergence}.}
\label{figure:TAarchitecture_basic}
\end{figure*}

Figure \ref{figure:TAarchitecture_basic} illustrates the structure of a TA with two actions and $2N$ states, where $N$ is the number of states for each action. When the TA is in any state between $0$ to $N-1$, the action ``Include" is selected. The action becomes ``Exclude" when the TA is in any state between $N$ to $2N-1$. The transitions among the states are triggered by a reward or a penalty that the TA receives from the environment, which, in this case, is determined by different types of feedback defined in the TM in the learning stage (to be explained later in this appendix).

\subsubsection{The operational concept of standard TM}


{\bf The input of a TM and its literals:} The \textit{input} of a standard TM is a one-dimensional Boolean feature vector: 
\[
F = [f_1, f_2, \cdots, f_o] \in \{0,1\}^o.
\]

The combined set of original features and their negations is referred to as \textit{literals}, and is defined as:
\begin{align}
L = &[f_1, f_2, \cdots, f_o, \neg f_1, \neg f_2, \cdots, \neg f_o] \nonumber\\
= &[l_1, l_2, \cdots, l_o, l_{o+1}, l_{o+2}, \cdots, l_{2o}] \in \{0,1\}^{2o},\nonumber
\end{align}
where $\neg f_1$ means the negation of the Boolean feature $f_1$. 

Formally, the literals are defined by:
\begin{equation}
\label{eqn:literals}
l_g = 
\begin{cases}
f_g, & \text{if } 1 \leq g \leq o; \\
\neg f_{g-o}, & \text{if } o+1 \leq g \leq 2o.
\end{cases}
\end{equation}

{\bf Clause constructions:} A \textit{clause} \( C^i_j \)\footnote{In this section, the index $i$ in $C_{j}^{i}$ denotes the class index, which is different from the layer index of $C_{j}^{i}$ in GraphTM.} is defined as a conjunction (AND operation) of included literals. A TM for recognizing class $i$ consists of $m$ conjunctive clauses, $C^i_{j}$, $j \in \{1,2,...,m\}$,  and is expressed as
\begin{equation}
\label{eqn:clause}
C^i_j = \left(\bigwedge\limits_{r \in \xi^i_j} l_r\right),
\end{equation}
where \( \xi^i_j \) denotes the set of indices of literals included in clause \( j \) for class \( i \). The inclusion or exclusion of each literal in the clause is controlled by its corresponding TA. In more detail, each literal in this clause has a corresponding TA, which determines whether the literal should be included or excluded from a clause depending on its current state (0 to $N-1$, include. $N$ to $2N-1$, exclude). For an input \( F \) with $o$ features, the total number of TAs in a clause is \( 2o \), matching the number of literals. For a TM with $m$ clauses, the total number of TAs is $m\times 2o$.

The output of the clause for an input $F$, $C_{j}^{i}(F)$, is obtained by the conjunction of included literals for the input $F$, as represented in Eq.~\eqref{eqn:clause}, and $C_{j}^{i}(F)\in\{0, 1\}$.   

The clauses for a certain class $i$ have polarities. Specifically, positive polarity clauses are odd-indexed, i.e., $C_{j}^{i}, j\in {1,3,...}$ while negative polarity clauses are even-indexed, i.e., $C_{j}^{i}, j\in\{2,4,...\}$. The positive clauses learn the features that belong to the class $i$ (i.e., the features with label 1 for class $i$) while negative clauses learn the features that does not belong to class $i$ (i.e., the features with label 0 for class $i$).

{\bf Classification:} The final classification result, for a certain input $F$,  is based on majority voting,  as
\begin{equation}
\label{eqn:cluase_op}
y = u  \left(\sum_{j = 1, 3, \cdots }^{m-1} C_{j}^i(F) -                     \sum_{j = 2, 4, \cdots}^{m} C_{j}^i(F) \right),
\end{equation}
where $u(\nu) = 1 \; \textbf{if} \;  \nu\geq0 \; \textbf{else} \; 0$. Here we assume $m$ is an even number. 

{\bf Example:} For a 2 bit input  $F=[f_1,f_2]$, its literals are $\{f_1, f_2,  \neg f_1, \neg f_2\}$.  Let's say if the TM has four clauses in the following forms after training: $C_1 = f_1 \land \neg f_2$, $C_2 = f_1 \land  f_2$, $C_3 = \neg f_1 \land  f_2$, $C_4 = \neg f_1 \land \neg f_2$, then the TM captures the XOR logic.

{\bf Training: } A standard TM learns in supervised manner, processing one input sample \((F, y)\) at a time. Since the TAs within each clause play a central role in shaping the behavior of the TM, the overall learning process is driven by the actions of these TAs. As described in earlier, each TA learns to choose its action, whether to include or exclude a literal, based on the feedback (rewards or penalties) it receives from the environment. Consequently, the effectiveness of TA learning, particularly in deciding the inclusion or exclusion of literals, is highly dependent on the quality of this feedback. 

This reward/penalty is provided through two types of feedback mechanisms, namely \textit{Type~I} and \textit{Type~II}. During training, each clause receives either Type~I or Type~II feedback for a given input sample, which in turn influences the team of TAs associated with that clause. This feedback determines whether each TA should receive a reward or a penalty.

More specifically, clauses update their corresponding TAs based on three key factors: the clause output value, the value of the literal within the clause, and the action taken by the associated TA (i.e., whether to \textit{include} or \textit{exclude} the literal). Each TA is guided by Type~I and Type~II feedback, as defined in Tables~\ref{table:type_i_feedback_1} and~\ref{table:type_ii_feedback_1}, respectively.

For clauses with \textbf{positive polarity} (i.e., learning features associated with label 1), Type~I feedback is applied when the training sample has a positive label \( y = 1 \), while Type~II feedback is triggered when \( y = 0 \). The hyperparameter \( s \) governs the granularity of clause specialization, with larger values promoting finer patterns and smaller values enabling broader generalization. Entries marked as "NA" in the tables indicate cases where feedback is not applicable.
For \textbf{negative polarity} clauses (i.e., those learning features associated with label 0), the same logic applies in reverse.

Since clauses in a TM are designed to capture different sub-patterns of the class, it is essential that they learn diverse sub-patterns rather than converging on the same one and overlooking others. To promote such diversity in learning, a user-defined hyper-parameter \( T \) is introduced. This parameter controls the feedback mechanism based on the target summation:

\[
\nu(F) = \sum_{\substack{j\; \text{is odd}}} C_{j}^{i}(F) - \sum_{\substack{j\; \text{is even}}} C_{j}^{i}(F),
\]

which influences the probability of clauses receiving Type~I or Type~II feedback. 
The probability of giving  Type~I feedback to clause $C_{j}^{i}(F)$ is computed as follows:

\[
\frac{T - \text{clip}(\nu(F), -T, T)}{2T}, \quad \text{when } j  ~\text{is odd and } y = 1,
\]
\[
\frac{T + \text{clip}(\nu(F), -T, T)}{2T}, \quad \text{when } j\text{ is even and } y = 0.
\]

Here, the \texttt{clip} function ensures that the value of \( \nu(F) \) remains within the range \([-T, T]\).

Similarly, the probability of assigning Type~II feedback is given by:

\[
\frac{T + \text{clip}(\nu(F), -T, T)}{2T}, \quad \text{when } j \text{ is even and } y = 1,
\]
\[
\frac{T - \text{clip}(\nu(F), -T, T)}{2T}, \quad \text{when } j \text{ is odd and } y = 0.
\]

This dynamic feedback allocation mechanism helps ensure that the TM learns a broader and more balanced set of sub-patterns by distributing learning effort more effectively across clauses.

\begin{table}[h!]
\centering
\caption{Type I Feedback --- Feedback upon receiving a sample with label $y=1$
~\citep{granmo2018tsetlin}.}
\begin{small}
\begin{tabular}{l c c c c c }
\toprule
\multicolumn{2}{r}{{\it Value of the clause} $C^i_j(F)$ }&\multicolumn{2}{c}{\True}&\multicolumn{2}{c}{\False}\\ 
\multicolumn{2}{r}{{\it Value of the Literal} $x_k$/$\lnot x_k$}&{\True}&{\False}&{\True}&{\False}\\
\midrule
\multirow{3}{*}{\bf Include Literal}&\multicolumn{1}{c}{$P(\mathrm{Reward})$}&$\frac{s-1}{s}$&NA&$0$&$0$\\
&\multicolumn{1}{c}{$P(\mathrm{Inaction})$}&$\frac{1}{s}$&NA&$\frac{s-1}{s}$&$\frac{s-1}{s}$\\
&\multicolumn{1}{c}{$P(\mathrm{Penalty})$}&$0$&NA&$\frac{1}{s} $&$\frac{1}{s}$\\
\addlinespace[5pt]
\multirow{3}{*}{\bf Exclude Literal }&\multicolumn{1}{c}{$P(\mathrm{Reward})$}&$0$&$\frac{1}{s}$&$\frac{1}{s}$ &$\frac{1}{s}$\\
&\multicolumn{1}{c}{$P(\mathrm{Inaction})$}&$\frac{1}{s}$&$\frac{s-1}{s}$&$\frac{s-1}{s}$ &$\frac{s-1}{s}$\\
&\multicolumn{1}{c}{$P(\mathrm{Penalty})$}&$\frac{s-1}{s}$&$0$&$0$&$0$\\
\bottomrule
\end{tabular}
\end{small}
\label{table:type_i_feedback_1}
\end{table}

\begin{table}[h!]
\centering
\caption{Type II Feedback --- Feedback upon receiving a sample with label $y=0$
~\citep{granmo2018tsetlin}.}
\begin{small}
\begin{tabular}{l c c c c c }
\toprule
\multicolumn{2}{r}{\it Value of the clause $C^i_j(F)$}&\multicolumn{2}{c}{\True}&\multicolumn{2}{c}{\False}\\ 
\multicolumn{2}{r}{\it Value of the Literal $x_k/\neg x_k$}&{\True}&{\False}&{\True}&{\False}\\
\midrule
\multirow{3}{*}{\bf Include Literal }&\multicolumn{1}{c}{$P(\mathrm{Reward})$}&$0$&$\mathrm{NA}$&$0$&$0$\\
&\multicolumn{1}{c}{$P(\mathrm{Inaction})$}&$1.0$&$\mathrm{NA}$&$1.0$&$1.0$\\
&\multicolumn{1}{c}{$P(\mathrm{Penalty})$}&$0$&$\mathrm{NA}$&$0$&$0$\\
\addlinespace[5pt]
\multirow{3}{*}{\bf Exclude Literal }&\multicolumn{1}{c}{$P(\mathrm{Reward})$}&$0$&$0$&$0$&$0$\\
&\multicolumn{1}{c}{$P(\mathrm{Inaction})$}&$1.0$&$0$&$1.0$ &$1.0$\\
&\multicolumn{1}{c}{$P(\mathrm{Penalty})$}&$0$&$1.0$&$0$&$0$\\
\bottomrule
\end{tabular}
\end{small}
\label{table:type_ii_feedback_1}
\end{table}

\subsection{Coalesced TM}
\label{app:CoTM}
GraphTM is based on Coalesced TM, which improves upon the standard Tsetlin Machine by introducing a concept of clause pool, where clauses are allowed to be shared across multiple classes, with each clause having class-specific weights. This enables more efficient use of clause resources and reduces memory consumption, while still maintaining class-specific decision capabilities.

Algorithm \ref{alg:decentralized_clause_updating} pseudocode demonstrates how clauses are updated in Coalesced TM. The core idea of the Coalesced Tsetlin Machine update step is to adjust the clause behaviors and their associated weights based on a voting mechanism and probabilistic feedback. For each class, the model computes a vote score by summing the outputs of all clauses, weighted by their respective class-specific weights. This score is clipped within a predefined range $[-T, T]$, where $T$ determines the granularity of the representation (larger $T$ and more clauses provide a finer range of vote sums). Then, an error signal is computed depending on whether the current class is the true label or not. Based on this error, each clause is probabilistically selected for an update. If selected, the model determines whether to apply Type I Feedback (to reinforce recognition of the correct class) or Type II Feedback (to suppress misclassification), depending on the alignment between the class label and the clause's voting polarity. This process incrementally refines the clause structure, enabling the machine to better distinguish between classes over time.

\begin{algorithm}[]
\scriptsize
\caption{Coalesced Tsetlin Machine Update Step for Multi-Label Classification}
\label{alg:decentralized_clause_updating}

\begin{algorithmic}[1]  
\STATE \textbf{Input:} Number of Classes $m$, Number of Clauses $n$, Multi-label Example $(\mathbf{x}, \mathbf{y}=[y_1, \ldots, y_m])$, Clauses $C_j \in \mathcal{C}$, Weights $w_{ij} \in \mathbf{W}$, Voting Target $T$, Pattern Specificity $s$
\STATE \textbf{Output:} Updated Clauses $\mathcal{C}'$, Updated Weights $\mathbf{W}'$, Class Prediction $\hat{\mathbf{y}}$

\STATE \textbf{Procedure:} CoalescedTMUpdate($X, \mathcal{C}, \mathbf{W}, P, b, T, s$)

\FOR{$i \gets 1$ to $m$} 
    \STATE $v_i \gets \mathbf{clip}\left(\sum_{j=1}^n w_{ij} C_j(\mathbf{x}), -T, T \right)$
    \STATE $e \gets T - v_i$ if $y_i = 1$ else $T + v_i$
    \STATE $p \gets w_{ij} \ge 0$
    \FOR{$j \gets 1$ to $n$}
        \IF{rand() $\le \frac{e}{2T}$}
            \IF{$y_i$ xor $p$}
                \IF{rand() $\le \frac{1}{m-1}$}
                    \STATE $C_j' \gets$ TypeIIFeedback($\mathbf{x}, C_j$)
                    \IF{$C_j(\mathbf{x})$}
                        \STATE $w_{ij}' \gets w_{ij} - (1 - 2p)$
                    \ENDIF
                \ENDIF
            \ELSE
                \STATE $C_j' \gets$ TypeIFeedback($\mathbf{x}, C_j, s$)
                \IF{$C_j(\mathbf{x})$}
                    \STATE $w_{ij}' \gets w_{ij} + (1 - 2p)$
                \ENDIF
            \ENDIF
        \ENDIF
    \ENDFOR
    \STATE $\hat{y}_i \gets v_i \ge 0$
\ENDFOR

\STATE \textbf{EndProcedure}
\end{algorithmic}

\end{algorithm}

\subsection{GraphTM}
\label{app:GraphTM}

{\bf Definitions and notations of the \ac{GraphTM}}
\begin{itemize}
    \item Graph: the input of a \ac{GraphTM} is a graph, which consists of nodes $X_n$ and edges $e$. An edge defines the relationship between two nodes. 
    \item Layers: A \ac{GraphTM} consists of multiple layers. $D$ is the number of layers in a \ac{GraphTM}. Layer $0$ is also called the node layer, Layers $1$ to $D-1$ are message layers.
    
    \item Hypervectors and feature bits: In the \ac{GraphTM}, features are represented by hypervectors. A hypervector is a sequence of binary bits, each of which can be used to represent a feature, and is referred to as a feature bit. A hypervector can be of different sizes. For example, the hypervecter $[0000~1111]$\footnote{The space in the middle is only for visually separating the hypervector into two parts; in reality, the hypervector does not contain this space.} has a size of 4, with its length being twice the size. The first half of a hypervector represents the original features, while the second half encodes their negations. In this example, $[0000~1111]$ indicates that no features are present in the hypervector. In contrast, the hypervector $[0100~1011]$ contains a single feature, with the active feature bit located at index 1 with a 0-based indexing.

    \item Node Symbols and $I_{sb}$: We use the term symbols to collectively denote node properties (node layer), messages (message layers) and edge types. A symbol at node level can be viewed as the smallest unit of input data features, for instance, a pixel of an image in an image processing problem, or a token of an input sentence in an NLP task. Each node symbol carries certain features, and properties of a node in the graph may incorporate features from multiple node symbols. For example, a node representing a $2 \times 2$ image patch would consist of 4 node symbols. 
    
    In the \ac{GraphTM} framework, each node symbol is assigned its own feature bit(s), with the indices specified by the vector $I_{sb}$. For example, if each node symbol occupies one feature bit, and their indices are given by $I_{sb} = [0, 1, 2, 3]$, this means that each node symbol corresponds to the 0th, 1st, 2nd, and 3rd bit of the hypervector, respectively. 
    
    Assuming the hypervector has a size of 8, the features from these four pixels can be encoded into a hypervector as $H = [11110000~00001111]$. This hypervector $H$ then characterizes the node that represents the $2 \times 2$ image patch.
    
    The length of $I_{sb}$ equals the number of node symbols ($N_{sb}$) multiplied by the number of feature bits per node symbol ($NB_{sb}$). The size of the hypervector needs to be large enough to ensure that different node symbols are assigned distinct feature bits if you want to avoid feature conflict.

    \item $C_j$: The $j$th clause in the \ac{GraphTM}. A \ac{GraphTM} can maintain multiple clauses, each trained to capture a data subpattern, i.e., a feature or combination of features of the data. A clause in a \ac{GraphTM} is constructed as a conjunction of clause components across different layers, i.e., $C_j = C^0_j \wedge C^1_j \wedge \dots \wedge C^{D-2}_j \wedge C^{D-1}_j$, where $C^i_j$ denotes the clause component in layer $i$ of the $j$-th clause. In a later example, we assume that only one clause is configured in the \ac{GraphTM}. Therefore, the clause is denoted as $C = C^0 \wedge C^1 \wedge \dots \wedge C^{D-2} \wedge C^{D-1}$, with the index $j$ omitted for simplicity.
    \item $H^0_n$: the node hypervector representing the features of node $X_n$, where the superscript $0$ indicates that the hypervector belongs to the node layer (Layer 0). The clause component $C^0$ is used to evaluate the node hypervector $H^0_n$, meaning that during training, $C^0$ learns the features encoded in $H^0_n$. In the testing phase, $H^0_n$ is compared with $C^0$ to determine whether it contains the subpattern represented by $C^0$. 
    
    \item Clause\footnote{Please note that the clause in this subsection is equivalent to a clause component $C^i$ from a specific layer in the \ac{GraphTM}, rather than the overall clause $C=C^0 \wedge C^1 \wedge \dots \wedge C^{D-2} \wedge C^{D-1}$.}-feature matching: clauses are trained to learn various subpatterns from the data. Each clause is a conjunction of a series of literals. In the \ac{GraphTM}, each literal is a bit in a hypervector. During training, the \ac{GraphTM} adjusts the clauses so that they converge to specific subpatterns. For example, consider the hypervector $H_0 = [0100~1011]$. We aim to train clauses to learn the feature represented by $H_0$. After training, a clause may converge to a form such as $C_0 = \neg b_0 \wedge b_1 \wedge \neg b_2 \wedge \neg b_3 \wedge b_4 \wedge \neg b_5 \wedge b_6 \wedge b_7$, where $b_i, i=0, 1,...,7$, is the $i$th bit in the input hypervector. At this point, if we compare each bit of $H_0$ with $C_0$, we find that $C_0(H_0) = \neg 0 \wedge 1 \wedge \neg 0 \wedge \neg 0 \wedge 1 \wedge \neg 0 \wedge 1 \wedge 1=True$, we thus can say that $C_0$ and $H_0$ match, and that the clause $C_0$ has successfully captured the features of $H_0$. 

    If $I_{sb} = [1, 2]$, specifying that symbols $P_0$ and $P_1$ (consider them as two pixels) correspond to the 1st and 2nd bits in a node hypervector respectively, then the hypervector $H_0=[0100~1011]$ contains feature of $P_0$, while $H_1=[0110~1001]$ contains features of both $P_0$ and $P_1$. The above clause that matches $H_0=[0100~1011]$, i.e., $C_0 = \neg b_0 \wedge b_1 \wedge \neg b_2 \wedge \neg b_3 \wedge b_4 \wedge \neg b_5 \wedge b_6 \wedge b_7$, can be expressed simply as $C_0 = P_0$, explicitly indicating that $C_0$ has captured a subpattern consisting only of features from $P_0$. Similarly, a clause that matches $H_1=[0110~1001]$, such as $C_1 = \neg b_0 \wedge b_1 \wedge b_2 \wedge \neg b_3 \wedge b_4 \wedge \neg b_5 \wedge \neg b_6 \wedge b_7$, can be written as $C_1 = P_0 \wedge P_1$, indicating that $C_1$ captures a subpattern consisting of features from both $P_0$ and $P_1$.
    
    If these clauses $C_0$ and $C_1$ are the results of \ac{GraphTM} training, and we later evaluate test input $H_2 = [1000~0111]$ with both clauses, we get: $C_0(H_2) = \neg 1 \wedge 0 \wedge \neg 0 \wedge \neg 0 \wedge 0 \wedge \neg 1 \wedge 1 \wedge 1 = False$, and $C_1(H_2) = \neg 1 \wedge 0 \wedge 0 \wedge \neg 0 \wedge 0 \wedge \neg 1 \wedge \neg 1 \wedge 1 = False$. This indicates that neither clause matches $H_3$. In other words, the input does not contain the subpatterns captured by $C_0$ or $C_1$. 

    Not every literal needs to be included in a clause. When a literal is omitted from a clause, it means the feature it represents is irrelevant. A clause can even be empty, denoted as $C = \phi$, indicating that no literals are included, that is, no features are relevant. During testing, an empty clause matches any input $H$, meaning $C(H)$ always evaluates to $True$.
        
    \item $M_n^0$: clause-feature matching result at node $X_n$ in Layer 0, denoted as $M_n^0=C^0(H^0_n)$. $M_n^0 = True$ indicates that $C^0$ matches $H^0_n$, while $M_n^0 = False$ means that $C^0$ and $H^0_n$ do not match. During training, $M_n^0 = True$ signifies that the clause component $C^0$ has successfully captured the node features of $X_n$. During testing, $M_n^0 = True$ indicates that node $X_n$ contains the features captured by clause component $C^0$.
    
    Note that if there are more than one clause in the system, then each clause corresponds to a clause-feature matching result in the node layer, denoted as $M^0_{jn}=C^0_j(H^0_n)$, where $j$ is the index of the clause. During training, it is possible for multiple clauses to learn the same subpattern. To prevent this redundancy, we typically introduce a hyperparameter $T$ to control the maximum number of clauses allowed to capture the same subpattern. 
    
    \item $H^i_n$, $i\in\{1,2,\ldots D-1\}$: the message hypervector at node $X_n$ in layer $i$. $i>0$ means the hypervectors are in message layers. During training, the clause component $C^i$, $i\in\{1,2,\ldots D-1\}$ is trained to learn the features represented by the message hypervector $H^i_n$; while in testing, $H^i_n$ is compared with $C^i$, to decide if $H^i_n$ contains the subpattern represented by $C^i$.
    
    \item $M_n^i$, $i\in\{1,2,\ldots D-1\}$: The overall clause-feature matching result at node $X_n$ in message layer $i$. $M_n^i=C^i(H^i_n) \wedge M_n^{i-1} =C^i(H^i_n) \wedge C^{i-1}(H^{i-1}_n) \wedge \ldots \wedge C^0(H^0_n)$, representing the conjunction of the matching result at layer $i$ and those from all preceding layers up to layer $i-1$. In other words, $M_n^i = 1$ indicates that $C^i$ matches with $H^i_n$ across all layers up to layer $i$. If the matching result is $False$ at any layer, then $M_n^i = False$. 

    Again, if there are more than one clause in the system, then each clause corresponds to a clause-feature matching result in the message layers, denoted as $M^i_{jn}= C^i_j(H^i_{n}) \wedge M_{jn}^{i-1} =C^i_j(H^i_{n}) \wedge C^{i-1}_j(H^{i-1}_n) \wedge ... \wedge C^0_j(H^0_n)$, where $j$ is the index of the clause. 

    \item Message bits, clause symbols and $I_{cl}$: 
    Clause symbols are symbols in the message layers, analogous to node symbols in the node layer. A clause symbol contains one or more message bits in a message hypervector. While feature bits describe the features of a node symbol, message bits represent the messages carried by a clause symbol. A message essentially conveys which clause matches the input. $I_{cl}$ is a vector specifying the indices of message bits for each clause. The length of $I_{cl}$ equals the number of clauses ($N_{cl}$) multiplied by the number of message bits per clause ($NB_{cl}$).
    
    \item $HV_{nd}$: node hypervector size. We define the size of the hypervector half of its length. This is because the second half of the hypervector is always the negation of the first half, which is redundant. For example, the length of the hypervector [1100~0011] is 8, and its size is 4. 
    \item $HV_{msg}$: message hypervector size. Similar to node hypervector size.
\end{itemize}


{\bf The sequence classification example:}\\
We use a simple sequence classification example to further illustrate the relevant definitions, notations, and working mechanisms of the \ac{GraphTM}. The example is to train a \ac{GraphTM} so that it can detect whether an input sequence of letters contains three consecutive ``A''s. We assume that each input sequence consists of five letters.

{\bf The input layer:} In the \ac{GraphTM}, each input sequence is encoded into a graph, which consists of multiple nodes and edges. In this example, each letter in the sequence is a node, denoted as $X_n,~n \in\{0, 1, 2, ..., 4\}$. The edges define the relationship between neighboring nodes. There are two types of edges in this graph: $e_{n,n+1}, ~n \in\{0,1,...,3\}$ denotes the left edge (from the perspective of the destination node) going from $X_n$ to its right neighbor (from the perspective of the source node) $X_{n+1}$; while $e_{n,n-1}, ~n \in\{1,2,...,4\}$ denotes the right edge going from node $X_n$ to the left neighbor $X_{n-1}$. We assign distinct integer values to each edge type to differentiate between them. In this example, $e_{n,n-1} = 1$ represents a right edge, while $e_{n,n+1} = 0$ represents a left edge.

In the \ac{GraphTM}, each node $X_n$ maintains a node hypervector $H^0_n$ which represents the property (feature(s)) of that node. 
In this example, the hypervector size is set to 8 bits, i.e., $HV_{nd} = 8$. Initially, $H^0_n = [00000000~11111111]$, where the first 8 bits are set to 0 and the second 8 bits are set to 1. The hypervector is doubled in size because the \ac{GraphTM} learns not only the original features (represented by the first 8 bits), but also their negations (represented by the second 8 bits). Initializing the first 8 bits with 0 indicates that no original features are present in the node at the beginning. The second half is always the negation of the first half, which is why it is entirely filled with 1s.

In this example, a symbol is a letter, and it is set to occupy two out of the 8 bits in the hypervector, i.e., $NB_{sb}=2$. We use different bits to represent different symbols, which we call feature bits of a symbol. For example, the 0th and 1st bits can be used as feature bits to represent ``A'', the 2nd and 3rd bits can be the feature bits of ``B'', and so on. 

Our sequence classification example aims to detect three consecutive ``A''s, making ``A'' the only symbol of interest. As a result, $I_{sb}$ contains only the indices of the two feature bits assigned to the symbol ``A''. Let’s assume $I_{sb} = [0, 1]$, meaning the 0th and 1st bits in the node hypervector correspond to the features of ``A''.

Recall that each node hypervector is initially set to $H^0_n=[00000000~11111111]$, indicating no features are present at the beginning. Now consider a sequence "BAAAE". For node $X_1 = A$, the corresponding hypervector $H^0_1$ will be updated to reflect the features of ``A'', resulting in $H^0_1 = [11000000~00111111]$. The same update occurs for $H^0_2$ and $H^0_3$, which correspond to nodes $X_2 = A$ and $X_3 = A$, respectively.

In contrast, for nodes $X_0 = B$ and $X_4 = E$, the corresponding hypervectors $H^0_0$ and $H^0_4$ remain unchanged from their initial state, as no feature bits have been allocated to these letters. Allocating feature bits to other letters is certainly possible, but it is not necessary in this context since they are not relevant to the learning objective.

The above process encodes a sequence of five letters into a graph consisting of five nodes and their corresponding edges. As shown in the input layer of Table~\ref{table:GTM_example1}, node features are represented by their corresponding hypervectors $H^0_n$, while edges are specified as either left (0) or right (1) type. The graph is the input of the \ac{GraphTM}.

{\bf The node and message layers:} A \ac{GraphTM} can have multiple layers, the number of layers is a hyper-parameter, denoted as $D$. Table~\ref{table:GTM_example1} shows a \ac{GraphTM} with 3 layers. A \ac{GraphTM} usually have a set of clauses configured, each of which is composed of clause components corresponding to different layers. For the sake of simplicity, we assume there is only one clause in the \ac{GraphTM}, denoted as $C=C^0 \wedge C^1 \wedge ... \wedge C^{D-2} \wedge C^{D-1}$. As shown in Table~\ref{table:GTM_example1}, in each layer $i$, the clause component $C^i$ learns the features (either node features or messages features) existing in that layer. $C^i(H^i_n)$ denotes the clause-feature matching results in layer $i$. Layer 0 of a \ac{GraphTM} is called the node layer, as it deals with the input node features represented by node hypervectors $H^0_n$. Layers $i, ~i \in \{1,2,...,D-1\}$ are the message layers, which handle message features.

Each message layer maintains its message hypervectors $H^i_n$, where $i \in \{1, 2, ..., D-1\}$, to represent the messages received by node $X_n$ in Layer $i$, primarily originating from neighboring nodes. These messages include which clause was triggered (i.e., when the clause feature match) in the previous layer and whether the message comes from the left or right neighbor (edge type). Assuming the message hypervector size is also 8, i.e., $HV_{msg} = 8$, then similar to the node hypervectors $H^0_n$, the message hypervectors $H^i_n$ are initialized as $[00000000~11111111]$, where the first 8 bits are set to 0 and the second 8 bits to 1, indicating that no message features are present at the beginning.

In our example, we assume $NB_{cl} = 2$. Since we have assumed that there is only one clause in the \ac{GraphTM}, $I_{cl}$ contains only the indices of the two feature bits associated with this clause. Let’s further assume that $I_{cl} = [4, 5]$, meaning the clause occupies the 4th and 5th bits in the message hypervector. These assumptions will be used to illustrate the message passing and updating process in the following sections.

{\bf Clause evaluation in Layer 0 and message passing in Layer 1:} When the graph is input to the \ac{GraphTM}, in Layer 0, the clause component $C^0$ is evaluated against each node hypervector $H^0_n$, represented as $M^0_n=C^0(H^0_n)$. The evaluation result is either $False$ or $True$. $M^0_n=True$ means that the property of $X_n$ matches the sub-pattern represented by $C^0$, while $M^0_n = False$ means they do not match.

Taking Node $X_2$ as an example, the evaluation is denoted as $C^0(H^0_2)$. As $H^0_2$ represents features of the symbol ``A'', if $C^0=A$, then clause and node property match, $C^0(H^0_2)=True$. 

A node $X_n$ will pass the message $M^0_n = True$\footnote{Recall that each clause component $C_j^0$ corresponds to a clause-feature matching result $M^0_{jn}$. Therefore, $M^0_{jn}$ indicates which clause matches with the node feature $H^0_n$. As there is only one clause in the system, the index of the clause is by default $0$, or omitted for simplicity.} to its neighbor(s). Messages with $M^0_n = False$ are not propagated as they do not convey useful information.


{\bf Message hypervector udpating:} We examine Node $X_2$ at Layer 1, whose neighbors are $X_1$ and $X_3$. Assuming the trained clause is $C_0=A$, this yields $M^0_1=C^0(X_1)=$True and $M^0_3=C^0(X_3)=$True. Then in Layer 1, the initial $H^1_2 = [00000000~11111111]$ is updated based on three factors: $M^0_1=1$, $M^0_3=1$ (indicating the clause index\footnote{there is only one clause in this example, so the index is omitted}); the edge type ($e_{1,2}=0$, $e_{3,2}=1$), and the indices of the clause message bits specified in $I_{cl}$ (i.e., $[4,5]$). The updating process follows Algorithm~\ref{alg:clause_feature_bit_update}.

\begin{algorithm}[]
\scriptsize
\caption{Message Hypervector Update for Node $X_2$ in Layer One}
\label{alg:clause_feature_bit_update}

\begin{algorithmic}[1]
\STATE \textbf{Input:} \\
Initial message hypervector $H^1_2 = [00000000~11111111]$; \\
Clause message bit indices $I_{cl} = [4,5]$; \\
Messages from neighbors: $M^0_1 = 1$ via $e_{1,2}$, $M^0_3 = 1$ via $e_{3,2}$.
\STATE \textbf{Output:} \\
Updated message hypervector $H^1_2$

\FOR{each neighbor $X_k$}
    \IF{$M^0_k = 1$}
        \STATE Retrieve clause message bit indices from $I_{cl}$ 
        \COMMENT{$[4,5]$ in this example}
        \STATE Retrieve edge type: $e_{k,2}$
        \COMMENT{$e_{k,2} \in \{0,1\}$}
        \STATE Compute new indices: $I_{\text{new}} \gets I_{cl} + [e_{k,2}, e_{k,2}]$
        \COMMENT{$I_{\text{new}} = [4,5]$ or $[5,6]$}
        \STATE Set $H^1_2[I_{\text{new}}] \gets 1$
        \STATE Set negating bits $H^1_2[8 + I_{\text{new}}] \gets 0$
    \ENDIF
\ENDFOR
\end{algorithmic}
\end{algorithm}

Briefly going through the updating process: $M^0_1 = 1$ is passed through edge $e_{1,2} = 0$, hence the message bits to be updated are \(I_{cl} + [e_{1,2}, e_{1,2}] = [4+0, 5+0] = [4, 5] \), i.e., the 4th and the 5th message bits in $H^1_2$. These bits are set to 1, and the corresponding negating message bits with index of $[8+4, 8+5]$ are set to 0, resulting in the updated $H^1_2 = [00001100~11110011]$. Through this update, the message hypervector $H^1_2$ encodes the information that the node property of the left neighbor $X_1=A$ matches the clause component $C^0$).

For the message $M^0_3 = 1$ coming from edge $e_{3,2} = 1$, The message bits indices to be updated are $I_{cl} + [e_{3,2}, e_{3,2}] = [5, 6]$. The message hypervector $H^1_2$ is further updated by setting the corresponding message bits to 1 and their negations to 0, resulting in $H^1_2= [00001110~11110001]$. This update further encodes that the node property of the right neighbor $X_3=A$ matches the clause component $C^0$. 

The same message passing and updating process applies to the remaining nodes. Table \ref{table:GTM_BAAAE} presents the updated message hypervectors for all nodes in Layer 1. Each updated message hypervector $H^1_n$ encodes which neighbor (as indicated by the edge type) in the previous layer possesses what features (as indicated by the matched clause indices).

{\bf Clause evaluation in Layer 1:} 
The clause component $C^1$ in Layer 1 is evaluated against the updated message hypervectors $H^1_n$ in this layer. For example, at node $X_2$, the evaluation is denoted as $C^1(H^1_2)$, which is further conjunctively combined with the result from the previous layer, i.e., $M^0_2 = C^0(H^0_2)$, to form the final clause-feature matching result in Layer 1: $M^1_2= M^0_2 \wedge C^1(H^1_2) = C^0(H^0_2)\wedge C^1(H^1_2)$.

As the information can be decoded from the message hypervector $H^1_2$ about which neighbor(s) (via edge type) possess(es) what features (via matched clause(s)), a clause component $C^1$ that matches $H^1_2$ can be written as $C^1 = l1:0 \wedge r1:0$. The number $1$ before the column sign is the index of the current message layer. The number $0$ after the column sign is the index of the matched clause\footnote{There is only one clause configured, the clause index is $0$, or simply omitted.}. $l$ and $r$ represent left and right edges, respectively.  

${\color{blue}l}{\color{green}1}:{\color{orange}0}$ basically means that the clause component in the previous layer $C_{\color{orange}0}^{\color{green}0}$ matches the node property of the {\color{blue}right} neighbor (message comes along the {\color{blue}} edge). Similarly, ${\color{blue}r}{\color{green}1}:{\color{orange}0}$ tells that the clause component $C_{\color{orange}0}^{\color{green}0}$ matches the node property of the {\color{blue}left} neighbor (message comes from the {\color{blue}right} edge). Therefore, $C^1=l1:0~ \wedge ~r1:0$ means $C_0^0$ matches node properties at both the right and left neighbors.


The above evaluation and message passing process applies to all nodes, producing two outputs in Layer 1: (1) Message hypervectors $H^1_n$, which, if updated, encode which neighbors possess what features in Layer 0. (2) Clause-feature matching results $M^1_n = C^0(H^0_n) \wedge C^1(H^1_n)$, which combines two components: $C^0(H^0_n)$, representing the evaluation between the clause and the node's own feature, and $C^1(H^1_n)$, representing the evaluation between the clause and the messages received from neighbors.

Table \ref{table:GTM_BAAAE} shows the detailed computation results of hypervectors in a two-layer \ac{GraphTM}, illustrating the encoded input into a graph and the encoded messages in the message layer.


{\bf Message passing and message hypervector updating in Layer 2:} When a clause matches at node $X_n$ in both Layer 1 and Layer 0, i.e., $M^1_n=1$, node $X_n$ passes a message to its neighbor(s) in Layer 2. Taking node $X_2$ as an example: if $M^1_1 = 1$, $X_2$ receives this message from $X_1$ along edge $e_{1,2} = 0$. Similarly, if $M^1_3 = 1$, $X_2$ also receives this message from $X_3$ via edge $e_{3,2} = 1$. These messages are used to update the message hypervector $H^2_2$ in Layer 2, following the same updating process as in Layer 1.

\textbf{The receptive field:} Since $H^1_1$ and $H^1_3$ may themselves contain messages passed from their respective neighbors ($X_0$ and $X_2$, and $X_2$ and $X_4$), the updated message hypervector $H^2_2$ may ultimately encode information propagated from $X_0$, $X_1$, $X_2$, $X_3$, and $X_4$. This implies that the receptive field of $X_2$ in Layer 2 includes neighbors up to two hops away. More generally, as the number of layers increases, the receptive field expands accordingly. This hierarchical structure is illustrated in Fig.~\ref{fig:receptivefiled}.


{\bf Clause evaluation in Layer 2:} The updated $H^2_2$ is evaluated against the clause component $C^2$ in Layer 2. The evaluation result, $C^2(H^2_2)$, is conjunctively combined with the result from the previous layer, $M^1_2$, to produce the final clause-feature matching result for Layer 2: $M^2_2=C^0(H^0_2) \wedge C^1(H^1_2) \wedge C^2(H^2_2)$. This result consists of three components: the node feature matching $C^0(H^0_2)$ from Layer 0 (capturing features of the node itself), the message feature matching $C^1(H^1_2)$ from Layer 1 (capturing immediate neighbors), and the message feature matching $C^2(H^2_2)$ from Layer 2 (capturing neighbors two hops away).

The same message passing and updating process applies to all other nodes, producing two types of outputs in Layer 2: the updated message hypervectors $H^2_n$, and the clause-feature matching results $M^2_n=C^0(H^0_n) \wedge C^1(H^1_n) \wedge C^2(H^2_n)$.

In general, a matched clause component in Layer $i,~i\in\{1,2,...,D-1\}$, can be written in the form $C^i=li:j \wedge ri:j \wedge ...$, indicating which neighbor (via left or right edges) contains what feature (via clause index $j$) in the previous layer $i-1$. 

{\bf The \ac{GraphTM} as a whole:} The overall clause $C$ in the \ac{GraphTM} is expressed as $C = C^0 \wedge C^1 \wedge C^2$. Each component of $C$ is evaluated on different hypervectors: the node hypervectors $H^0_n$ in the node layer, and the message hypervectors $H^i_n$ in the corresponding message layers. 

In our example, we simplified the scenario by assuming the presence of only a single clause in the system. In general, however, multiple clauses are typically employed, with each clause undergoing the full message passing and updating process described above. The only difference is that the vector $I_{cl}$ will be longer, and each message hypervector may encode messages from multiple clauses.

\textbf{Sparsity in the hypervector space:} From the message hypervector updating process, it can be seen that the size of the message hypervector should not only ensure that different clauses are assigned distinct message bits, but also maintain sparsity in the space. This is particularly important when many edge types are present, as they tend to occupy bit positions adjacent to the clause feature bits. A sparse space helps avoid potential conflicts between messages.

{\bf Classification:} The evaluation of the clause on the entire input graph in our sequence classification example is the disjunctive combination of the clause evaluation from all nodes in the last layer, which can be written as $M = M^2_0 \vee M^2_1 \vee M^2_2 \vee M^2_3 \vee M^2_4$. This means that if the clause feature match across all layers, at any one node, then $M=\text{True}$, i.e., the input graph matches the clause. If this is the only one clause in the \ac{GraphTM}, the match result of this clause is also the final classification of the \ac{GraphTM}, which indicates a positive classification.

When there are more than one clause in the \ac{GraphTM}, 
The evaluation result of each clause, i.e., $M_j$\footnote{$M_j=M^2_{j0} \vee M^2_{j1} \vee M^2_{j2} \vee M^2_{j3} \vee M^2_{j4}$ in our example.}, can be combined using a weighted sum, or by voting, to achieve the final classification.

At this point, we have thoroughly explained how input can be encoded into a graph, how \ac{GraphTM} passes messages between nodes, how message hypervectors are updated, and how the final classification is determined. The explanation is based on the evaluation process. However, \ac{GraphTM}'s training is also straightforward to understand. Given a sufficient number of labeled samples, the clauses in \ac{GraphTM} learn both the node features and the message features at each layer through the designed mechanism. The learning process relies on each TA (Tsetlin Automaton), which, under the coordination of the TM's feedback tables, adjusts the contribution of each bit of feature by including or excluding the corresponding literal from the clauses. Ultimately, the clauses converge towards sub-patterns that are beneficial for classification.

{\bf Results Analysis:} Please refer to \ref{sec:seq} for the detailed analysis on the experiment results.

{\bf Another Experiment:} We conducted another experiment classifying sequences into three classes based on the number of consecutive ``A''s they contain, corresponding to “A”, “AA”, and “AAA”, respectively. The \ac{GraphTM} consists of 3 layers, and the number of clauses was set to 4. The training solution is: 
\begin{itemize}
    \item $C_0= A \wedge r1:1 \wedge r1:2 \wedge r2:1$; $[-6, 8, -2]$
    \item $C_1= l1:0 \wedge l1:2 \wedge l2:0 \wedge l2:1$; $[0, -8, 6]$
    \item $C_2= A \wedge l2:0 \wedge l2:1$; $[-1, -3, 1]$
    \item $C_3= \neg A$; $[3, -3, -5]$
\end{itemize}

Table \ref{table:exp2clause} shows the clauses in a structured manner (columns 2, 3, 4). It also shows the clause evaluation on Node $X_n$ (columns 5, 6 and 7), with the messages captured by $C^1_j$ and $C^2_j$ being traced back to the node layer (columns 6 and 7). 

Taking $C_0$ as an example, according to Table \ref{table:exp2clause}, the clause evaluation at node $X_n$ can be expressed as:
\begin{tiny}
\begin{align}
\label{eq:C_0exp2}
    C_0(X_n) & = C_0^0(X_n) \wedge C_0^1(X_n) \wedge C_0^2(X_n) \\
    & = C_0^0(X_n) \wedge \left(C_1^0(X_{n-1}) \wedge C_2^0(X_{n-1})\right) \wedge C_1^1(X_{n-1}) \nonumber\\
    & = C_0^0(X_n) \wedge \left(C_1^0(X_{n-1}) \wedge C_2^0(X_{n-1})\right) \wedge C_2^0(X_{n}) \nonumber\\
    & = \mathcal{M}(\neg A, X_n) \wedge \left(\mathcal{M}(\phi, X_{n-1}) \wedge \mathcal{M}(A, X_{n-1})\right) \wedge \mathcal{M}(A, X_{n}) \nonumber\\
    & = \mathcal{M}(\neg A, X_n) \wedge \mathcal{M}(A, X_{n-1}) \wedge \mathcal{M}(A, X_{n}) \nonumber
\end{align}
\end{tiny}

Eq.~\ref{eq:C_0exp2} is derived in the same manner as Eq.~\ref{eq:C_0}, with the only difference lying in Layer 2. We now explain how $C_0^2(X_n)$ is traced back to the node layer. Since $C_0^2 = r2{:}1$, this indicates that a message about $C_1$ was sent along a right edge in Layer 2, implying that the left neighbor of $X_n$ matches the clause component $C^1_1$, denoted as $C^1_1(X_{n-1})$. As $C^1_1 = l1{:}2$, this further indicates that a message about $C_2$ was sent along a left edge in Layer 1, implying that the right neighbor of $X_{n-1}$ matches the clause component $C^0_2$, denoted as $C^0_2(X_n)$. In this way, $C_0^2(X_n)$ is traced back to the node layer as $C^0_2(X_n)$.

Similarly, we have the other clause evaluation at node $X_n$, they are:
\begin{tiny}
\begin{align}
\label{eq:C_1exp2}
    C_1(X_n) = & C_1^0(X_n) \wedge C_1^1(X_n) \wedge C_2^1(X_n) \\
    = & C_1^0(X_n) \wedge C_2^0(X_{n+1}) \wedge C_0^1(X_{n+1}) \wedge C_1^1(X_{n+1}) \nonumber\\
    = & C_1^0(X_n) \wedge C_2^0(X_{n+1}) \wedge C_1^0(X_{n}) \wedge C_2^0(X_{n}) \wedge C_2^0(X_{n+2}) \nonumber\\
    = & C_1^0(X_n) \wedge C_2^0(X_{n}) \wedge C_2^0(X_{n+1}) \wedge C_2^0(X_{n+2}) \nonumber\\
    = & \mathcal{M}(\phi, X_n) \wedge \mathcal{M}(A, X_{n}) \wedge \mathcal{M}(A, X_{n+1}) \wedge \mathcal{M}(A, X_{n+2}), \nonumber\\
    = & \mathcal{M}(A, X_{n}) \wedge \mathcal{M}(A, X_{n+1}) \wedge \mathcal{M}(A, X_{n+2}), \nonumber
\end{align}
\end{tiny}

\begin{tiny}
\begin{align}
\label{eq:C_2exp2}
    C_2(X_n) = & C_2^0(X_n) \wedge C_2^1(X_n) \wedge C_2^2(X_n) \\
    = & C_2^0(X_n) \wedge C_0^1(X_{n+1}) \wedge C_1^1(X_{n+1}) \nonumber\\
    = & C_2^0(X_n) \wedge C_1^0(X_{n}) \wedge C_2^0(X_{n}) \wedge C_2^0(X_{n+2}) \nonumber\\
    = & C_1^0(X_{n}) \wedge C_2^0(X_{n}) \wedge C_2^0(X_{n+2}) \nonumber\\
    = & \mathcal{M}(\phi, X_{n}) \wedge \mathcal{M}(A, X_{n}) \wedge \mathcal{M}(A, X_{n+2}), \nonumber\\
    = & \mathcal{M}(A, X_{n}) \wedge \mathcal{M}(A, X_{n+2}), \nonumber
\end{align}
\end{tiny}

\begin{tiny}
\begin{align}
\label{eq:C_3exp2}
    C_3(X_n) = & C_3^0(X_n) \wedge C_3^1(X_n) \wedge C_3^1(X_n) \\
    = & C_3^0(X_n) \nonumber\\
    = & \mathcal{M}( \neg A, X_n) \nonumber
\end{align}
\end{tiny}

Assuming that the test sequence ``BBAEE'' enters to the \ac{GraphTM}, by applying Eqs. \ref{eq:C_0exp2} - \ref{eq:C_3exp2}, we obtain Table \ref{table:clause_on_nodeexp2}, which displays the evaluation results of each clause on each node, and reveals that the activated clause is $C_3$. The weighted sum is thus [3, -3, -5], indicating Class 0.

The 3-layer \ac{GraphTM} thus correctly identifies that the sequence ``BBAEE'' contains only one occurrence of ``A''.

\begin{table}[htbp]
\centering
\caption{Clause evaluation results at each node, when the input sequence is ``BBAEE''.}
\begin{tabular}{cccccc}
\toprule
$C_j(X_n)$ & $X_0$=B & $X_1$=B & $X_2$=A & $X_3$=E & $X_4$=E \\
\midrule
$C_0$ & $False$ & $False$ & $False$ & $False$ & $False$ \\
$C_1$ & $False$ & $False$ & $False$ & $False$ & $False$ \\
$C_2$ & $False$ & $False$ & $False$ & $False$  & $False$ \\
$C_3$ & $True$ & $True$ & $False$ & $True$ & $True$ \\
\bottomrule
\end{tabular}
\label{table:clause_on_nodeexp2}
\end{table}

\begin{table*}[htbp]
\centering
\caption{Clause components and clause traceability in a 3-layer \ac{GraphTM} for Experiment 2.}
\begin{small}
\begin{tabular}{r|rrr|rrr|r}
\toprule
\textbf{$C_j$} & \textbf{$C_j^0$} & \textbf{$C_j^1$} & \textbf{$C_j^2$} & \textbf{$C_j^0(X_n)$} & \textbf{$C_j^1(X_n)$ to Layer 0} & \textbf{$C_j^2(X_n)$ to layers 1 and 0} & \textbf{weights} \\
\midrule
$C_0$ & A & r1:1 & r2:1 & $C_0^0(X_n)$ & $C_1^0(X_{n-1})$ & $C_1^1(X_{n-1})$ $\rightarrow$  $C_2^0(X_n)$ & [-6, 8, 2] \\
      &   & r1:2 &      &             & $C_2^0(X_{n-1})$ &                  &                               \\
\midrule
$C_1$ & $\phi$ & l1:2 & l2:0 & $C_1^0(X_n)$ & $C_2^0(X_{n+1})$ & $C_0^1(X_{n+1})$ $\rightarrow$  $C_1^0(X_n)$ & [0, -8, 6] \\
      &        &      &      &              &                  &                   $C_2^0(X_n)$ &                            \\
      &        &      & l2:1 &              &                   & $C_1^1(X_{n+1})$ $\rightarrow$  $C_2^0(X_{n+2})$ & \\
\midrule
$C_2$ & A & $\phi$ & l2:0 & $C_2^0(X_n)$ & $\phi$ & $C_0^1(X_{n+1})$ $\rightarrow$  $C_1^0(X_n)$ & [-1, -3, 1] \\
      &   &        &      &              &        &                   $C_2^0(X_n)$  &              \\
      &   &        & l2:1 &              &        & $C_1^1(X_{n+1})$ $\rightarrow$  $C_2^0(X_{n+2})$ & \\
\midrule
$C_3$ & $\neg$ A & $\phi$ & $\phi$ & $C_3^0(X_n)$ & $\phi$ & $\phi$ $\rightarrow$  $\phi$ & [3, -3, -5] \\
\bottomrule
\end{tabular}
\end{small}
\label{table:exp2clause}
\end{table*}




\newpage
\begin{landscape} 
\begin{table}[!ht]
    \centering
    \caption{A three-layer \ac{GraphTM} with the input graph containing five nodes}
    \label{table:GTM_example1}
    {\fontsize{8pt}{10pt}\selectfont
    \begin{tiny}
    \begin{tabular}{ccccccc}
        \toprule
        \multirow{2}{*}{Input}  & Node hypervector: $H^0_n$ & $H^0_0$ & $H^0_1$ &$H^0_2$  &$H^0_3$  & $H^0_4$ \\
        & edge type: e & $e_{0,1} = 0 $ & $e_{1,0} = 1, ~ e_{1,2} = 0$ & $e_{2,1} = 1, ~ e_{2,3} = 0$ & $e_{3,2} = 1, ~ e_{3,4} = 0$ & $e_{4,3} = 1$ \\
        \midrule
         \multirow{2}{*}{Layer 0} & Clause component $C^0$ &  &  &  &  &  \\
         &Clause-node matching $M_n^0$  & $M_0^0= C^0(H_0^0)$ & $M_1^0= C^0(H_1^0)$&$M_2^0= C^0(H_2^0)$ &$M_3^0= C^0(H_3^0)$  & $M_4^0= C^0(H_4^0)$\\
        \midrule
        \multirow{4}{*}{Layer 1} & Feature bits index vector $I_{cl}$ & \multirow{2}{*}{$I_{cl},M_1^0,e_{1,0}\rightarrow H_0^1$} & \multirow{2}{*}{$I_{cl}, M_0^0,e_{0,1},M_2^0,e_{2,1}\rightarrow H_1^1$} & \multirow{2}{*}{$I_{cl}, M_1^0, e_{1,2}, M_3^0, e_{3,2}\rightarrow H_2^1$} & \multirow{2}{*}{$I_{cl}, M_2^0, e_{2,3}, M_4^0, e_{4,3}\rightarrow H_3^1$} & \multirow{2}{*}{$I_{cl}, M_3^0, e_{3,4} \rightarrow H_4^1$} \\
         &Updated msg-feature $H_n^1$  & & & & & \\
         & Clause component $C^1$& \multirow{2}{*}{$M_0^1= C^1(H_0^1)\wedge M_0^0$} & \multirow{2}{*}{$M_1^1= C^1(H_1^1)\wedge M_1^0$} & \multirow{2}{*}{$M_2^1= C^1(H_2^1)\wedge M_2^0$} & \multirow{2}{*}{$M_3^1= C^1(H_3^1)\wedge M_3^0$} & \multirow{2}{*}{$M_4^1= C^1(H_4^1)\wedge M_4^0$} \\
         &Clause-msg matching $M_n^1$ & & & & & \\
        \midrule
        \multirow{4}{*}{Layer 2} & Feature bits index vector $I_{cl}$ & \multirow{2}{*}{$I_{cl},M_1^1,e_{1,0}\rightarrow H_0^2$} & \multirow{2}{*}{$I_{cl}, M_0^1,e_{0,1},M_2^1,e_{2,1}\rightarrow H_1^2$} & \multirow{2}{*}{$I_{cl}, M_1^1, e_{1,2}, M_3^1, e_{3,2}\rightarrow H_2^2$} & \multirow{2}{*}{$I_{cl}, M_2^1, e_{2,3}, M_4^1, e_{4,3}\rightarrow H_3^2$} & \multirow{2}{*}{$I_{cl}, M_3^1, e_{3,4} \rightarrow H_4^2$} \\
         & Updated msg-feature $H_n^2$  & & & & & \\
         & Clause component $C^2$& \multirow{2}{*}{$M_0^2= C^2(H_0^2)\wedge M_0^1$} & \multirow{2}{*}{$M_1^2= C^2(H_1^2)\wedge M_1^1$} & \multirow{2}{*}{$M_2^2= C^2(H_2^2)\wedge M_2^1$} & \multirow{2}{*}{$M_3^2= C^2(H_3^2)\wedge M_3^1$} & \multirow{2}{*}{$M_4^2= C^2(H_4^2)\wedge M_4^1$} \\
         &Clause-msg matching $M_n^2$ & & & & & \\
        \bottomrule
    \end{tabular}
    \end{tiny}
    }
    
\end{table}

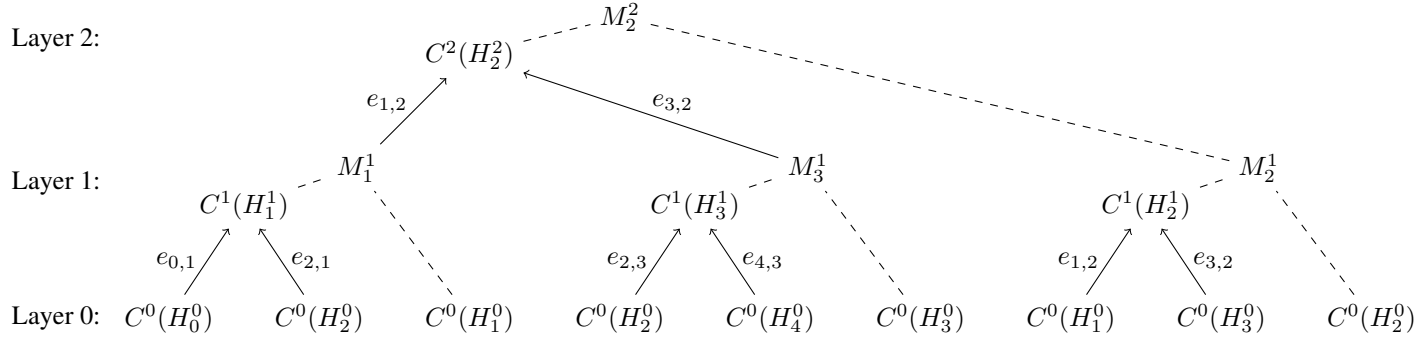
\begin{figure}[!ht]
\begin{center}
\begin{tikzpicture}

    \node[draw=none] (L0) at (-5.5,0) {Layer 0:};
    \node[draw=none] (H01) at (-4,0) {\(C^0(H_0^0)\)};
    \node[draw=none] (H02) at (-2,0) {\(C^0(H_2^0)\)};    
    \node[draw=none] (H03) at (0,0) {\(C^0(H_1^0)\)};
    \node[draw=none] (H04) at (2,0) {\(C^0(H_2^0)\)};
    \node[draw=none] (H05) at (4,0) {\(C^0(H_4^0)\)};
    \node[draw=none] (H06) at (6,0) {\(C^0(H_3^0)\)};
    \node[draw=none] (H07) at (8,0) {\(C^0(H_1^0)\)};
    \node[draw=none] (H08) at (10,0) {\(C^0(H_3^0)\)};
    \node[draw=none] (H09) at (12,0) {\(C^0(H_2^0)\)};

    \node[draw=none] (e01) at (-3.9,0.75) {$e_{0,1}$};
    \node[draw=none] (e02) at (-2.1,0.75) {$e_{2,1}$};    
    \node[draw=none] (e03) at (2.1,0.75) {$e_{2,3}$};
    \node[draw=none] (e04) at (3.9,0.75) {$e_{4,3}$};
    \node[draw=none] (e05) at (8.1,0.75) {$e_{1,2}$};
    \node[draw=none] (e06) at (9.9,0.75) {$e_{3,2}$};

    \node[draw=none] (L1) at (-5.5,1.8) {Layer 1:};
    \node[draw=none] (H111) at (-3,1.5) {\(C^1(H_1^1)\)};
    \node[draw=none] (H112) at (3,1.5) {\(C^1(H_3^1)\)};
    \node[draw=none] (H113) at (9,1.5) {\(C^1(H_2^1)\)};

    \node[draw=none] (H121) at (-1.5,2) {\(M_1^1\)};
    \node[draw=none] (H122) at (4.5,2) {\(M_3^1\)};
    \node[draw=none] (H123) at (10.5,2) {\(M_2^1\)};

    \node[draw=none] (e11) at (-1.1,2.85) {$e_{1,2}$};
    \node[draw=none] (e12) at (2.7,2.85) {$e_{3,2}$};

    \node[draw=none] (L2) at (-5.5,3.7) {Layer 2:};
    \node[draw=none] (H211) at (0,3.5) {\(C^2(H_2^2)\)};
    \node[draw=none] (H221) at (2,4) {\(M_2^2\)};

    \draw [->] (H01) -- (H111);
    \draw [->] (H02) -- (H111);
    \draw [dashed] (H111) -- (H121);
    \draw [dashed] (H03) -- (H121);
    \draw [->] (H04) -- (H112);
    \draw [->] (H05) -- (H112);
    \draw [dashed] (H112) -- (H122);
    \draw [dashed] (H06) -- (H122);
    \draw [->] (H07) -- (H113);
    \draw [->] (H08) -- (H113);
    \draw [dashed] (H113) -- (H123);
    \draw [dashed] (H09) -- (H123);

    \draw [->] (H121) -- (H211);
    \draw [->] (H122) -- (H211);
    
    \draw[dashed] (H123) -- (H221);
    \draw[dashed] (H211) -- (H221);
    
\end{tikzpicture}
\end{center}
\caption{Hierarchical message passing structure across layers, from the perspective of Node $X_2$. Solid arrows indicate potential information flow, while dashed lines represent conjunctive connections between components.}
\label{fig:receptivefiled}
\end{figure}

\begin{table}[!ht]
    \centering
    \caption{Hypervectors in a two-layer \ac{GraphTM}, with ``BAAAE'' as the input. $I_{sb}=[1,2]$, $I_{cl}=[4,5]$.}
    \label{table:GTM_BAAAE}
    {\fontsize{8pt}{10pt}\selectfont
    \begin{tiny}
    \begin{tabular}{cccccc}
        \toprule
        \multirow{3}{*}{Input Graph} &   $X_1$=B &$X_2$=A &$X_3$=A &$X_4$=A &$X_5$=E \\
        & $H^0_1=[00000000~11111111]$ &$H^0_2=[11000000~00111111]$  &$H^0_3=[11000000~00111111]$  & $H^0_4=[11000000~00111111]$ & $H^0_5=[00000000~11111111]$ \\
        & $e_{1, 2} = 0 $ & $e_{2, 1} = 1, ~ e_{2, 3} = 0$ & $e_{3, 2} = 1, ~ e_{3, 4} = 0$ & $e_{4, 3} = 1, ~ e_{4, 5} = 0$ & $e_{5, 4} = 1$ \\
        \midrule
        Layer 0 & $M^0_1=C^0(H^0_1)=0$ & $M^0_2=C^0(H^0_2)=1$ & $M^0_3=C^0(H^0_3)=1$ & $M^0_4=C^0(H^0_4)=1$ & $M^0_5=C^0(H^0_5)=0$ \\
        \midrule
        \multirow{3}{*}{Layer 1}
        & $H^1_1=[00000110~11111001]$ 
        & $H^1_2=[00000110~11111001]$ 
        & $H^1_3=[00001110~11110001]$ 
        & $H^1_4=[00001100~11110011]$ 
        & $H^1_5=[00001100~11110011]$\\
        &$M^1_1=C^1(H^1_1)\wedge C^0(H^0_1)$
        &$M^1_2=C^1(H^1_2)\wedge C^0(H^0_2)$
        &$M^1_3=C^1(H^1_3)\wedge C^0(H^0_3)$
        &$M^1_4=C^1(H^1_4)\wedge C^0(H^0_4)$
        &$M^1_5=C^1(H^1_5)\wedge C^0(H^0_5)$\\
        &$=0\wedge 0 = 0$
        &$=0\wedge 1 = 0$
        &$=1\wedge 1 = 1$
        &$=0\wedge 1 = 0$
        &$=0\wedge 1 = 0$\\
        \bottomrule
    \end{tabular}
    \end{tiny}
    }    
\end{table}
\end{landscape}

\subsection{GraphTM Hyperparameters}
\subsubsection{Disconnected Nodes}
The MNIST dataset consisted of 60,000 training images, and 10,000 test images. Similarly to MNIST, the F-MNIST consisted of 60,000 training images, and 10,000 test images. The CIFAR-10 dataset consisted of 50,000 training images, and 10,000 test images. The hyperparameters used for the three datasets are shown in Table~\ref{tab:hyperparameters_disconnected}.
\begin{table*}[htbp]
\caption{Hyperparameter values used for the \ac{GraphTM} and the \ac{CoTM} in the experiments on MNIST, Fashion-MNIST, and CIFAR-10.}
\centering
    \begin{tiny}
        \begin{tabular}{l c c c c c c c c c }
            \toprule
            Dataset & Clauses & Number of symbols & Convolution Window & Hypervector Size & Depth & Epochs & T & s & Max included literals \\
            \midrule
            MNIST & 2,500 & 138 & $10 \times 10$ & 128 & 1& 30 & 3,125 & 10.0 & - \\
            F-MNIST & 40,000 & 124 & $3 \times 3$ & 128 & 1 & 30 & 15,000 & 10.0 & -  \\
            CIFAR-10 & 80,000 &  2,500 & $8\times8$ & 128 & 1 & 30 & 15,000 & 20.0  & 32 \\
            \bottomrule
        \end{tabular}
    \end{tiny}
    \label{tab:hyperparameters_disconnected}
\end{table*}

\subsubsection{Connected Nodes with Superpixels}
The hyperparameters used for the experiments on the MNIST Superpixel dataset (Sec.~\ref{subsec_superpixels}) are listed in Table~\ref{tab:hyperparameters_superpixel}.
\begin{table*}[htbp]
\caption{Hyperparameter values used for the \ac{GraphTM} in the experiment on the MNIST Superpixel dataset.}
\centering
        \begin{tabular}{l c c c c c c c c }
            \toprule
            Clauses & Hypervector Size & Message Hypervector Size & Depth & Epochs & T & s & Max included literals \\
            \midrule
            60,000 & 128 & 256 & 35 & 50 & 100,000 & 0.5 & 32 \\
            \bottomrule
        \end{tabular}
        \label{tab:hyperparameters_superpixel}
\end{table*}

\subsubsection{Sentiment Polarity Classification}
The hyperparameters used for the experiments on the IMDB, Yelp, and MPQ datasets (Sec.~\ref{subsec:sentiment}) are listed in Table~\ref{tab:hyperparameters_IMDB}.
\begin{table*}[htbp]
\caption{Hyperparameter used for the \ac{GraphTM} and Standard \ac{TM} (StdTM) in the experiments on IMDB, Yelp, and MPQA.}
\centering
    \begin{tabular}{c c c c c c c c c } 
    \toprule
    Model & Clauses & T      & s  & Hypervector size & Depth & Epochs & Message Hypervector size \\
    \midrule
    GraphTM     & 10,000   & 100,000 & 15 & 2048             & 2     & 40     & 1024 \\
    StdTM & 10,000   & 100,000 & 15 & -                & -     & 40     & -\\
    \bottomrule
    \end{tabular}
    \label{tab:hyperparameters_IMDB}
\end{table*}

\subsubsection{Tracking Action Coreference}
The hyperparameters used for the experiments on the Tangram dataset (Sec.~\ref{subsec:actioncoreference}) are listed in Table~\ref{tab:hyperparameters_tangram}.
\begin{table*}[h!]
    \caption{Hyperparameter used for the \ac{GraphTM} for experiments on Tangram dataset.}
    \centering
    \begin{tabular}{c c c c c c c c c } 
    \toprule
    Utterances & Clauses & T      & s  & Hypervector size & Depth & Epochs & Message Hypervector size \\
    \midrule
    3  & 850   & 9,000 & 1 & 256               & 6     & 50    & 256\\\
    5 & 800   & 9,000 & 1 & 512               & 12     & 50    & 256\\
    \bottomrule
    \end{tabular}
    \label{tab:hyperparameters_tangram}
\end{table*}

\subsubsection{Recommendation Systems}
In this experiments the \ac{GCN} implemented with an input feature dimension of 64, a hidden dimension of 128, and an output dimension of 64. The model parameters were optimized using the Adam optimizer with a learning rate of 0.01. The \ac{GraphTM} employed 2000 clauses, a $T$ of 10,000, and $s$ of 10.0, utilizing hypervector encoding with a $HV\_size$ of 4096, $HV\_bits$ = 256, message size = 256 and message bits = 2 to capture graph relationships. For the standard \ac{TM}, we used 2,000 clauses with a maximum of 32 literals, $T$ set at 10,000, $s$ of 10.0. 

\subsubsection{Top-N Recommendation Systems}
The hyperparameters used on the Top-N MovieLens recommendation dataset (Sec. \ref{app:topn}) are listed in Table \ref{tab:movielens}.
\begin{table*}[htbp]
\caption{Hyperparameter values used for the \ac{GraphTM} in the experiment on the MovieLens Top-N recommendation dataset.}
\centering
        \begin{tabular}{l c c c c c c c c }
            \toprule
            Clauses & Hypervector Size & Hypervector Bits & Depth & Epochs & T & s & Max included literals \\
            \midrule
            4096 & 2096 & 64 & 1 & 10 & 500 & 5.0 & 256 \\
            \bottomrule
        \end{tabular}
        \label{tab:movielens}
\end{table*}

\subsubsection{Viral Genome Sequence Data}
All experiments were carried out for 10 epochs, except for those involving an increased sequence length, for which 20 epochs were used. The hyperparameters are shown in Table~\ref{tab:scalability_experiment}.

\begin{table*}[htbp]
    \caption{Hyperparameter values used for the scalability experiments on the viral genome sequence dataset with different amount classes.}
    \centering
    \begin{tiny}
        \begin{tabular}{l c c c c c c c c c}
            \toprule
            Classes & Clauses & T & s & Epochs & Max included literals & Number of symbols & Hypervector size & Message Hypervector Size & Depth   \\
            \midrule
             2  & 500 & 2,000 & 1 & 10 & 200 & 64 & 512 & 512 & 2\\
             3 & 700 & 2,000 & 1 & 10 & 200 & 64 & 512 & 512 & 2 \\
             4 & 1,000 & 2,000 & 1 & 10 & 200 & 64 & 512 & 512 & 2 \\
             5 & 2,000 & 2,000 & 1 & 10 & 200 & 64 & 512 & 512 & 2\\
            \bottomrule
        \end{tabular}
    \end{tiny}
    \label{tab:scalability_experiment}
\end{table*}

\subsubsection{Multivalue Noisy XOR}
The dataset consisted of 50,000 training samples, and 5,000 test samples. The hyperparameter values for the multivalue noisy XOR experiments are listed in Table \ref{tab_hyperparameters_XOR}.
\begin{table*}[h!]
    \caption{Hyperparameter values used for experiments on the noisy multivalue XOR dataset.}
    \centering
    \begin{tiny}
        \begin{tabular}{l c c c c c c c}
            \toprule
            Clauses & T & s & Number of symbols & Hypervector size & Message Hypervector Size & Depth  & Noise \\
            \midrule
            $\{4, 50, 200, 1000, 2000\}$ & $10 \times$clauses & 2.2 & 500 & 2048 & $\{2048, 4096, 8196\}$ & 2 & 1\%  \\
            \bottomrule
        \end{tabular}
    \end{tiny}
    \label{tab_hyperparameters_XOR}
\end{table*}

\subsection{Top-N Recommendation Systems} \label{app:topn}
\begin{table} [H] 
 \centering\begin{tabular}{c c c c}  \hline  Model & nDCG@10 & HR@10 & ItemCoverage \\ [0.5ex]  \hline
MostPop & 0.0364 & 0.2942 & 10 \\ [0.5ex] 
Random & 0.0052 & 0.0873 & \textbf{\textit{3582}} \\ [0.5ex] 
StdTM & 0.0487 & 0.3684 & 257 \\ [0.5ex] 
GraphTM & \textbf{0.1089} & \textbf{0.596} & 521 \\ [0.5ex] 
ItemKNN & 0.1083 & 0.5776 & 1231 \\ [0.5ex] 
MultiDAE & 0.0969 & 0.5576 & 1401 \\ [0.5ex] 
UserKNN & 0.1007 & 0.5336 & \textbf{1844} \\ [0.5ex] 
NeuMF & 0.0661 & 0.4444 & 111 \\ [0.5ex] 
BPRMF & 0.0828 & 0.4498 & 106 \\ [0.5ex] \hline
\end{tabular} 
 \caption{Performance metrics for the models on the MovieLens dataset. For ItemCoverage, the second best performance is marked in addition to the best performance. The ELLIOT implementation of the MostPop algorithm has been slightly modified to allow for repeat recommendations to reinforce the reader's intuition regarding the ItemCoverage metric.}
 \label{tab:movieLensResults}
 \end{table}

RS as a domain is a particularly opaque form of machine learning, driving the need to explore interpretable options. The main body of this paper has evaluated RS performance as a rating prediction problem, however, a more traditional approach within this domain is to predict future user-item interactions by ranking all items. This section discusses such a problem formulation. Our experiments evaluate the relative performance of the GraphTM as compared to standard TMs and a selection of well-established RS algorithms, including model-based (MultiDAE\cite{multidae}, NeuMF\cite{neumf}, BPRMF\cite{bprmf}), nearest neighbor-based (UserKNN\cite{userknn}, ItemKNN\cite{itemknn}), and heuristic (MostPop, Random). All experiments were conducted on the MovieLens dataset\cite{movielensinteractions}; 20\% of each user's interactions were hidden as a test dataset via temporal hold out. We treat movie ranking as a large, implicit, multi-label classification problem. Each of the items in the dataset are assigned a class, the top-n classes in terms of clause activations are evaluated as recommendations. As inputs, each users' train interaction history is formed as a sequence of movies, with their corresponding movie id and listed genre(s) included as features. Each item has an edge to the item temporally consumed before and after it.\\

The metrics used are Hit Rate at 10 (HR$@$10), normalized Discounted Cumulative Gains at 10 (nDCG$@$10), and Item Coverage. HR$@$10 assigns a score of 1 if at least one of the top-10 recommended items appears in the test set, the final score reflects the mean across all users.
nDCG$@$10 accounts for both hits and their ranking positions, giving higher weight to higher-ranked hits: $$DCG@k = \frac{1}{U}\sum_{u=1}^{U}\sum_{i=1}^{k}\frac{2^{rel_i} - 1}{log_2(i+ 1)},$$ 
$$nDCG@k =  \frac{DCG@k}{IDCG@k},$$ 
where $rel_i=1$ if the item at position $i$ is in the test set, and 0 otherwise. IDCG$@$k denotes the ideal DCG, i.e., the maximum possible score for the given parameters. ItemCoverage measures the total number of unique items recommended, providing a concise indication of the diversity of each algorithm’s recommendations. As offline RS evaluations are particularly prone to inconsistent results and difficult to reproduce, all evaluations are performed using ELLIOT, a Python module built to standardize recommendation benchmarking\cite{elliot}.\\

The results are shown in Table~\ref{tab:movieLensResults}. GraphTM outperforms all tested baselines, including the standard TM. Compared to its predecessor, GraphTM is particularly well suited to this RS formulation due to its ability to handle variable-length sequences. Due to the significant class imbalance, many model-based RS algorithms tend to converge to local optima that recommend only the most popular items, as reflected by their low ItemCoverage. While not achieving the same coverage as nearest-neighbor methods, GraphTM mitigates this popularity bias better than most model-based alternatives.





\end{document}